\documentclass[letterpaper, 10 pt, conference]{ieeeconf}  

\IEEEoverridecommandlockouts                              

\usepackage{graphics} 
\usepackage{graphicx}
\usepackage{amsmath} 
\usepackage{amssymb}  
\usepackage{cite}
\usepackage{tikz}
\usepackage{siunitx}
\usepackage{lipsum}
\usepackage{hyperref}
\usetikzlibrary{arrows.meta}
\usetikzlibrary{calc}

\usepackage{subcaption}

\renewcommand{\L}{\mathcal{L}}
\newcommand{\T}{\mathcal{T}}
\newcommand{\V}{\mathcal{V}}

\graphicspath{{Images/}}

\title{\LARGE \bf
A Pendulum-Driven Legless Rolling Jumping Robot }

\author{Jake Buzhardt$^{\dagger 1}$, Prashanth Chivkula$^{\dagger 1}$ and Phanindra Tallapragada$^{1}$
\thanks{$^{\dagger}$ Indicates equal contribution.}
\thanks{View the supplemental video at \url{https://youtu.be/9hKQilCpeaw}}
\thanks{$^{1}$The authors are with the Department of Mechanical Engineering, Clemson University, Clemson, SC, USA,
{\tt\small \{jbuzhar,pchivku,ptallap\}@clemson.edu}}
}

\begin{document}

\maketitle
\thispagestyle{empty}
\pagestyle{empty}

\begin{abstract}
In this paper, we present a novel rolling, jumping robot.  The robot consists of a driven pendulum mounted to a wheel in a compact, lightweight, 3D printed design.  
We show that by driving the pendulum to shift the robot's weight distribution, the robot is able to obtain significant rolling speed, achieve jumps of up to 2.5 body lengths vertically, and clear horizontal distances of over 6 body lengths.  The robot's dynamic model is derived and simulation results indicate that it is consistent with the rolling motion and jumping observed on the robot.  The ability to both roll and jump effectively using a minimalistic design makes this robot unique and could inspire the use of similar mechanisms on robots intended for applications in which agile locomotion on unstructured terrain is necessary, such as disaster response or planetary exploration. 
\end{abstract}

\section{Introduction}
The mobility of a robot in unstructured terrain can greatly improve with the ability to execute small hops or jumps repeatedly with little relaxation.  
Even in structured settings, such added agility can allow mobile robots to clear obstacles or climb up stairs. 
In recent years, many jumping robot designs have been proposed, with most of them achieving a jump using a spring-like mechanism to store potential energy \cite{msu_jumper_2013,penn_jerboa_2015,grillo_2007,salto_2016}. 
Such a mechanism can then be rapidly released to convert the potential energy to kinetic energy of a jump. 

In this paper, we take an alternative approach by proposing a rolling, jumping wheel robot which achieves both of these modes of motion by shifting its internal mass distribution without the aid of any elastic element. 
The design is inspired by the complex dynamics of a rolling hoop with an unbalanced mass, the so-called Littlewood hoop \cite{littlewood}, which has been the subject of many theoretical and experimental investigations \cite{tokieda_hopping_1997, moffatt_prs_2005, ivanov_rcd_2008, bronars_prs_2019, tallapragada_passive_2019}. 
Just as the original works on passive walking \cite{mcgeer1990passive,garcia1998simplest} inspired the development of many actuated walking mechanisms \cite{collins2005efficient}, here we let the passive rolling jumping mechanism of Littlewood’s hopping hoop serve as inspiration for an actuated rolling, jumping robot.  The specific design used in this paper is a wheel that contains an actuated internal pendulum. This wheel mechanism can roll due to a torque exerted via the internal pendulum and a high speed swing of the pendulum produces a discontinuity in the normal reaction on the robot from the ground leading to a jump.  With such a framework, even in the first design iteration, the robot is capable of a vertical jump height of over two body lengths and a horizontal jump distance of over 6 body lengths. Furthermore the robot can execute repeated jumps without a long relaxation time. The repeated jumping motion takes advantage of the slip motion of the wheel upon impact and the resulting friction force.

The proposed legless rolling-jumping robot is distinct in design and mechanics from the existing few jumping robots that can also roll or tip over repeatedly. Past rolling-jumping robots such as \cite{armour_bb_2007, shinichi_ICRA_2007, misu_iros_2018} still relied on elastic components or  shape memory to store and release energy. For instance the jollbot in \cite{armour_bb_2007} is an elastic spheroid which uses a pendulum mechanism for rolling, but can also compress and change shape significantly to store elastic potential energy that can be released for a jump of about $0.6$ body lengths. The two wheeled robot in \cite{misu_iros_2018} has an elastic chassis that can snap through and buckle and produce a jump that is $0.25$ body lengths in height and $1$ body lengths in width.  Despite the lack of any elastic components, the novel legless rolling-jumping robot presented in this paper can jump significantly higher and farther. 

Another class of related robots are those which employ momentum-based strategies for rolling, walking, or jumping.  For example, the robot in \cite{hayashi_high-performance_2001} is able to jump up and down stairs by swinging a pair of pendulum arms.  Several other works have recently proposed mechanisms which utilize the transfer of angular momentum that occurs from suddenly braking a rapidly spinning momentum wheel to initiate a jumping, tipping, or standing motion \cite{hockman2017design,ho2017mascot,geist_wheelbot_2022,romanishin20153d,muehlebach2016cubli}.  However, these robots are either unable to locomote aside from their jumping behavior or rely on other mechanisms for more efficient movement.  Many such jumping robots do not take advantage of other efficient modes of locomotion that may be possible on different terrains.
Similarly, several robots have used pendulum-based mechanisms for rolling or walking, such as pendulum-driven rolling spheres \cite{kayacan2012sphere} and torso-actuated walking mechanisms \cite{howell2000torsobiped,sanchez_differential_2020}, but these applications focus only on ground-based locomotion without consideration of how such a mechanism could be used for jumping. 
In this paper we propose a legless robot design that is capable of two modes of motion -- rolling and jumping. 
Even more importantly this paper presents a fresh look at the theoretical models and mechanics that can lead to the design of robots with superior mobility.  

\section{Robot design and system description} \label{sec:robot_design}

\begin{figure}
	    \centering
     \begin{tikzpicture}%
        \node[anchor=south west,inner sep=0] (f1) at (0,0) {\includegraphics[width=0.49\linewidth]{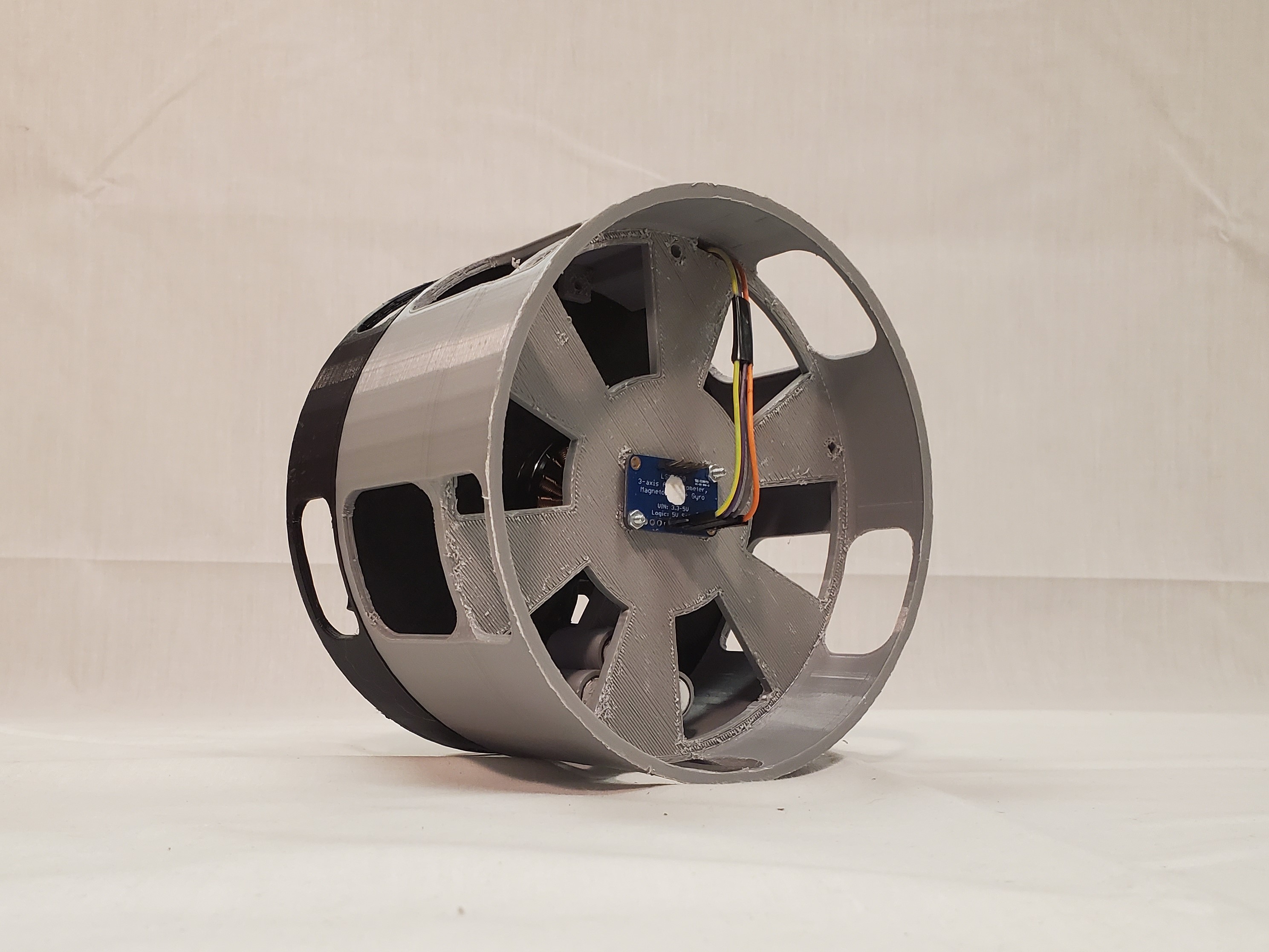}};
        \draw (f1.north east) rectangle (f1.south west);
        \node[circle,draw,minimum size = 0.4cm, inner sep=0pt] (A) at (3.4,2.1){\small d};
        \draw (A) --++ (-1.15,-0.55);
             

    \end{tikzpicture}
    \begin{tikzpicture}
        \node[anchor=south west,inner sep=0] (f2) at (0,0) {\includegraphics[width=0.49\linewidth]{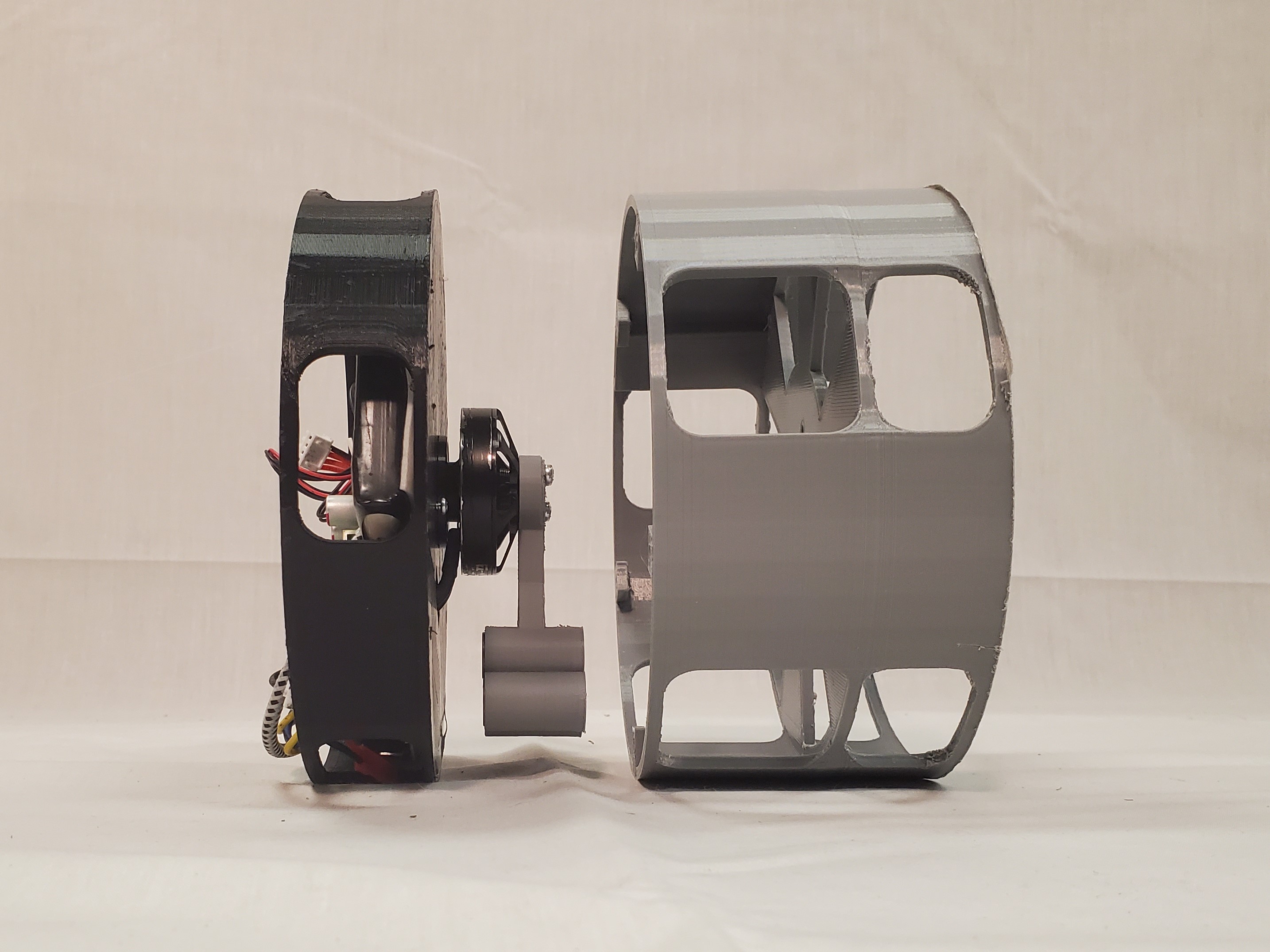}};
        \draw (f2.north east) rectangle (f2.south west);
    \end{tikzpicture}
    \\[1ex]
    \begin{tikzpicture}
        \node[anchor=south west,inner sep=0] (f3) at (0,0) {\includegraphics[width=0.49\linewidth]{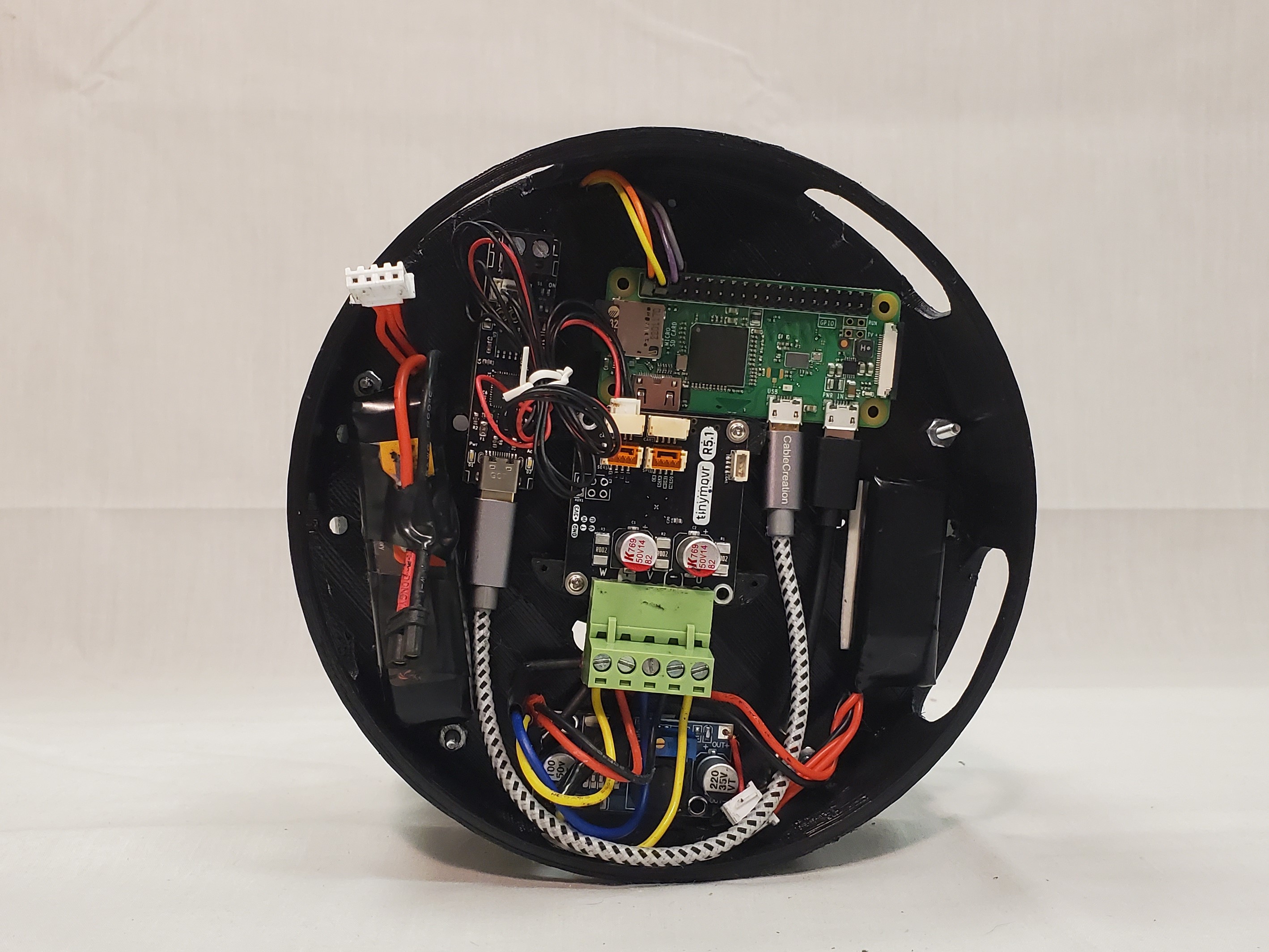}};
        \draw (f3.north east) rectangle (f3.south west);
        \node[circle,draw,minimum size = 0.4cm, inner sep=0pt] (A) at (3.9,2.5){\small b};
        \draw (A) --++ (-1.3,-0.5);
        \node[circle,draw,minimum size = 0.4cm, inner sep=0pt] (B) at (0.45,1.2){\small c};
        \draw (B) --++ (1.7,0.25);
        \node[circle,draw,minimum size = 0.4cm, inner sep=0pt] (E) at (1.2,2.85){\small e};
        \draw (E) --++ (0.5,-0.9);
    \end{tikzpicture}
    \begin{tikzpicture}
        \node[anchor=south west,inner sep=0] (f4) at (0,0) {\includegraphics[width=0.49\linewidth]{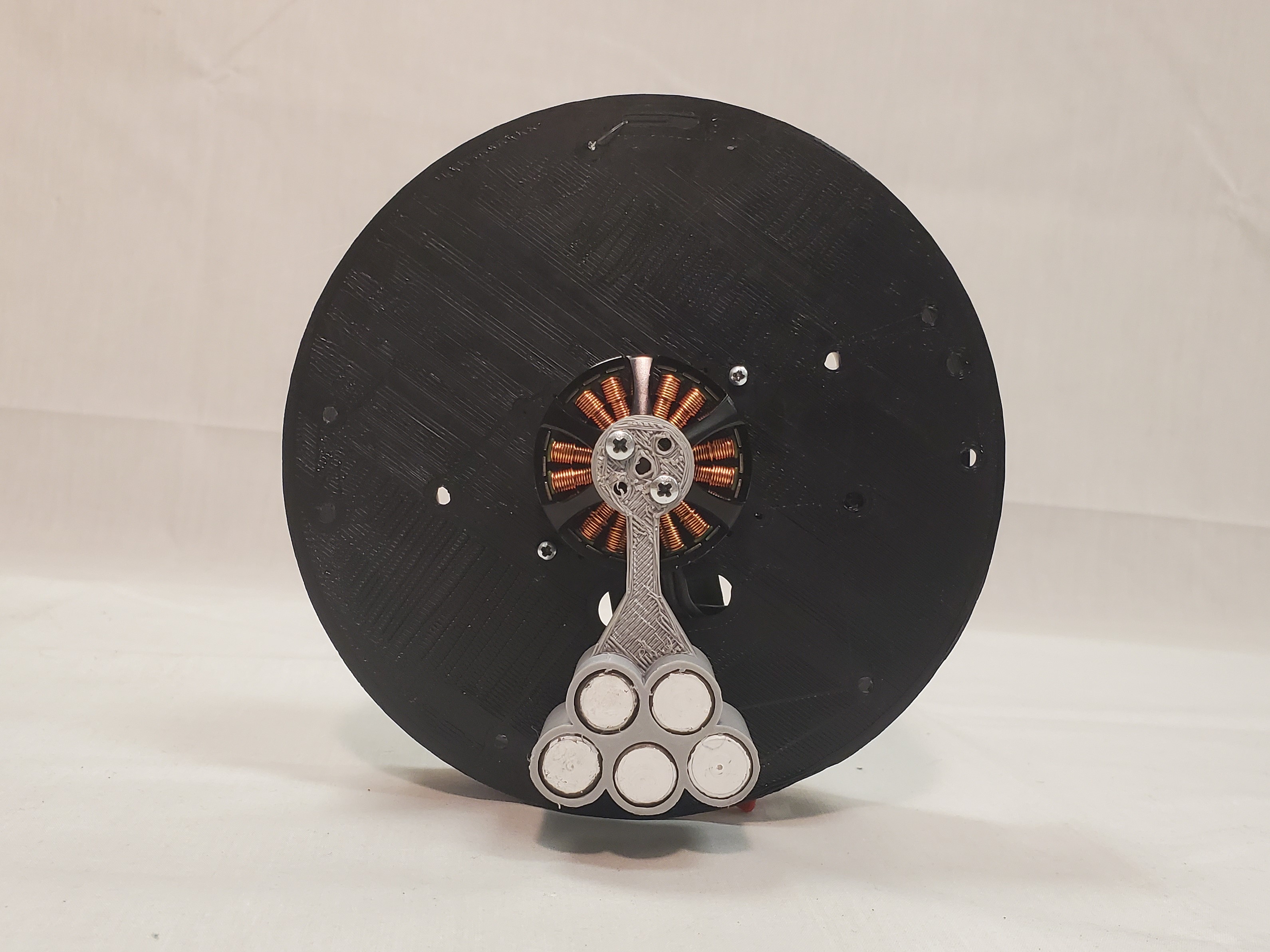}};
        \draw (f4.north east) rectangle (f4.south west);
        \node[circle,draw,minimum size = 0.4cm, inner sep=0pt] (A) at (3.9,2.5){\small a};
        \draw (A) --++ (-1.5,-0.7);
\end{tikzpicture}
	    \caption{Top left: pendulum-driven jumping wheel robot.  Top right: side view of disassembled robot.  Bottom right: Front view of pendulum mounted to a brushless DC motor affixed to the back hoop.  Bottom left: rear view of robot showing electronics. Labels correspond to labels in Table \ref{tab:components}.}
	\label{fig:robot_assembly}
	\end{figure}
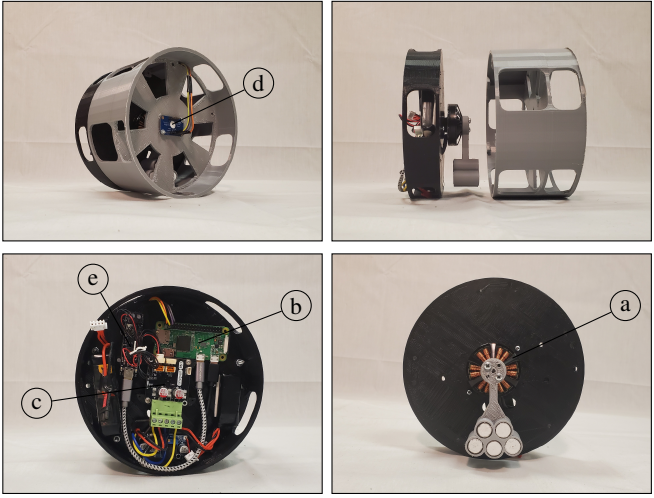

The jumping mechanism in consideration here requires a wheel body with an approximately uniform weight distribution with an offset mass attached to an arm which can be driven by a motor from the center of the wheel.  Such a robot was constructed from two 3D printed hoops with a pendulum driven by a brushless DC motor affixed to the back hoop. This setup is shown in Fig. \ref{fig:robot_assembly} and the key electronic components are summarized in Table \ref{tab:components}.

The total mass of the robot with all components attached is approximately 600 g and the offset mass of 125 g is attached to a pendulum which rotates at a radial distance of 51 mm from the geometric center of the hoop.  
The diameter of the hoop is 152 mm.  The length of the two hoops together is 106mm, with the front and rear hoops having lengths 75 mm and 31 mm.  These relative lengths are chosen so that the weight distribution of the robot is approximately centered along the axis passing through the centers of the hoops when the pendulum and motor are mounted to the back hoop. 
 \begin{table}[h]
     \centering
     \caption{Component specifications}
     \begin{tabular}{l|l|c}
     \hline
          Component & Specification & Label \\
     \hline
          Motor & T-motor MN4006 KV380 & (a)  \\
          Single Board Computer & Raspberry Pi Zero W & (b) \\
          Motor Driver & Tinymovr R5.1 & (c) \\
          IMU & Adafruit LSM9DS0 9 DOF & (d)\\
          CAN-USB Adapter & CANine USB-CAN adapter & (e)\\
          Batteries & 11.1 V (3S), 450 mAh\\
                    & 7.4 V (2S), 400 mAh\\
     \end{tabular}
     \label{tab:components}
 \end{table}

The electronics and controls configuration of the robot consists of a T-motor MN4006 380 Kv brushless DC motor (BLDC) driven by a Tinymovr R5.1 motor driver.  Being designed for use in drones, this BLDC has a very compact and lightweight design, with a diameter of 44mm and weight of 68g, while being capable of delivering a torque of up to 0.376Nm when used with the Tinymovr R5.1.  The Tinymovr driver uses field oriented control (FOC) to drive the BLDC using a built-in magnetic encoder to measure motor rotation and speed.  In our application, the Tinymovr is used in velocity-control mode, in which a PI controller feeds forward a motor current value to the FOC control loop based on errors from a motor velocity reference.  These control loops run at 20 kHz.  The Tinymovr communicates with a Raspberry Pi Zero W over a CAN connection via the CANine CAN-USB adapter.  The Raspberry Pi allows for higher level controllers to be implemented, as it delivers velocity setpoint values to the velocity control loop running on the Tinymovr.  
However, in this work, the velocity setpoint values are fed forward in an open-loop manner, as higher level control strategies for this robot are still the subject of ongoing work.

\section{Legless jumping - concept and model} \label{sec:model}

The mechanical system in consideration consists of a thin hoop of mass $m_o$ and radius $R$ with a mass moment of inertia $I_o$ about it's geometric center.  Attached to the hoop from the center is a pendulum of mass $m_p$ and length $l_p$.  The arm of the pendulum is taken to be massless and the pendulum is driven relative to the hoop by a torque applied by a motor connecting the pendulum to the hoop.  

\begin{figure}
    \centering
\begin{subfigure}[b]{0.53\linewidth}
        \includegraphics[width = \linewidth]{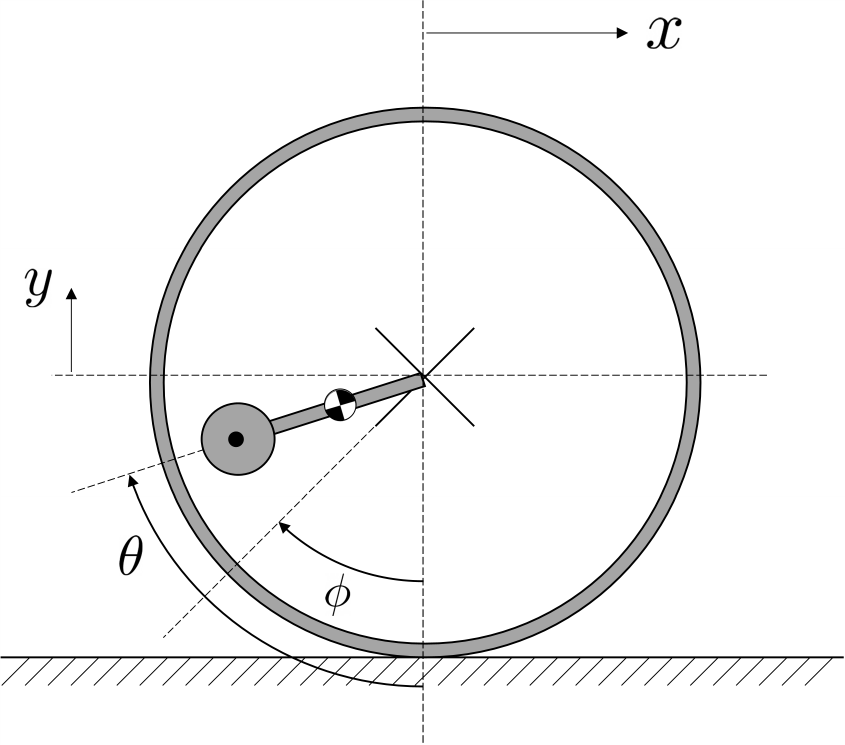}
        \caption{}
        \label{fig:hoop_schematic}
\end{subfigure}
\begin{subfigure}[b]{0.45\linewidth}
        \includegraphics[width=\linewidth]{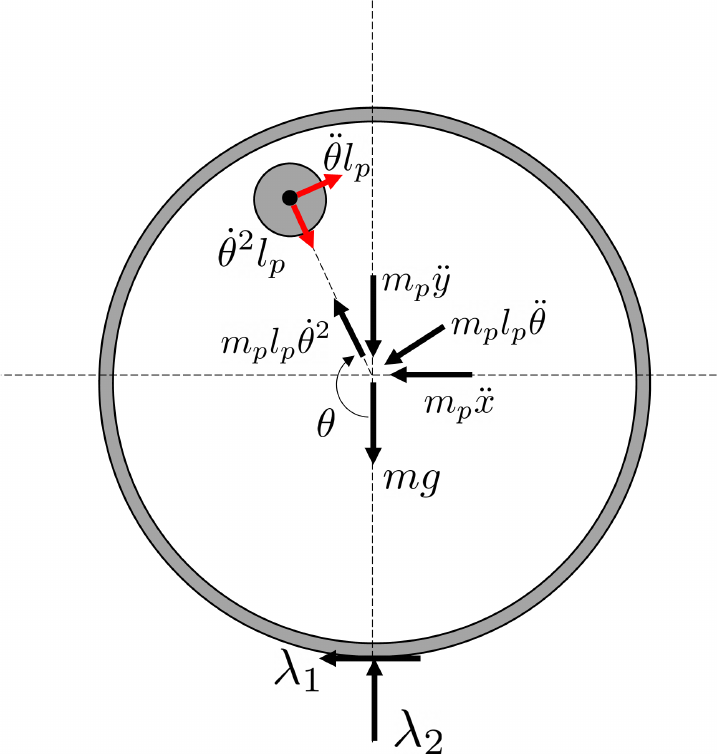}
        \caption{}
    \label{fig:hoop_fbd}
\end{subfigure}
\caption{(a) Pendulum-driven wheel robot system with generalized coordinates $\phi$, $\theta$, $x$, $y$. (b) Free body diagram of detached wheel, indicating the effects of the pendulum acceleration on the wheel.}
\end{figure}

We express the motion of the system in terms of the generalized coordinates $\mathbf{q} = [\phi, \theta, x, y]^T$ depicted in Fig. \ref{fig:hoop_schematic}. The angles $\phi$ and $\theta$ denote the angles of the hoop and the pendulum, respectively, both relative to a spatially fixed frame of reference.  The $\phi=\theta=0$ configuration corresponds to the pendulum in the vertically downward position with no relative angular displacement between the pendulum and the hoop. The $x$ and $y$ coordinates refer to the horizontal and vertical displacement of the geometric center of the hoop, respectively, relative to a spatially fixed frame.  The $y=0$ position is defined as the configuration in which the hoop is in contact with the ground.

 The kinetic energy of the system is 
\begin{equation}
    \T = \frac{1}{2}m_o|\mathbf{v}_o|^2 + \frac{1}{2}I_o\dot{\phi}^2 + \frac{1}{2}m_p|\mathbf{v}_p|^2
\end{equation}
where the velocities of the center of the hoop and the pendulum are 
$\mathbf{v}_o = \dot{x}\,\mathbf{i} + \dot{y}\,\mathbf{j}$ and $\mathbf{v}_p = \left(\dot{x} - l_p\dot{\theta}\cos\theta\right) \mathbf{i} + \left(\dot{y} + l_p\dot{\theta}\sin\theta\right) \mathbf{j}$, 
where $\mathbf{i}$ and $\mathbf{j}$ are the unit vectors associated with the spatially fixed reference frame. 
The potential energy of the system is 
\begin{equation}
    \V = m_ogy + m_pg(y-l_p\cos\theta)
\end{equation}
Then, taking the Lagrangian as $\L = \T - \V$, we can write the equations of motion from the Euler-Lagrange equations:
\begin{equation}
    \frac{d}{dt}\left(\frac{\partial \L}{\partial \dot{q}_i}\right) - \frac{\partial \L}{\partial q_i} = \sum_j\lambda_j\frac{\partial f^{\text{con}}_j}{\partial q_i} + f^{\text{ext}}_i
\end{equation}
for $i=1,\dots, 4$ where $f^{\text{con}}_j$ ($j=1,2$) are the holonomic constraints acting on the system, $\lambda_j$ are the constraint forces required to satisfy these constraints, and $f^{\text{ext}}_i$ are the non-conservative generalized forces acting on the system. 
The holonomic constraints associated with this system are those of pure rolling, $f^{\text{con}}_1 = \phi R-x = 0$, and contact with the flat surface, $f^{\text{con}}_2 = y = 0$.
The external generalized forces are those associated with the torque, $\tau$, applied by the motor to the pendulum and the hoop.

\subsection{Rolling phase}
During the rolling phase of motion, the wheel remains in contact with the ground and rolls without slip. 
Applying the Euler-Lagrange equation with both constraints active yield the following Equations of motion during the rolling phase.
\begin{subequations}\label{eq:rolling_eom}
    \begin{align}
    (I_o + mR^2)\ddot{\phi} + m_pl_pR(-\ddot{\theta}\cos\theta + \dot{\theta}^2\sin\theta) &= -\tau\\
     m_pl_p^2\ddot{\theta} + m_pl_p(-R\ddot{\phi}\cos\theta +g\sin\theta) &= \tau
    \end{align}
\end{subequations}
The expressions for the friction, $\lambda_1$, required to prevent slipping and the normal reaction, $\lambda_2$, during rolling are given as follows. 
\begin{subequations}
    \begin{align}
        \lambda_1 &= m_pl_p\left(-\dot{\theta}^2\sin\theta+\ddot{\theta}\cos\theta  \right) - mR\ddot{\phi} \\
        \lambda_2 &= m_pl_p\left(\dot{\theta}^2\cos\theta + \ddot{\theta}\sin\theta\right) + mg \label{eq:normal_reaction}
    \end{align}
\end{subequations}
The transition from a rolling phase to a flight phase occurs when the normal force vanishes and the velocity of the center of mass has a positive upward component. It is also possible for an intermediate phase to occur in which the friction provided by the rolling surface is insufficient to maintain the pure rolling constraint. In this case, the hoop could undergo slipping, skidding, or gliding motions \cite{taylor_dynamics_2010,bronars_prs_2019}.  In this work, we will assume that such effects are negligible.  

\subsection{Flight phase}
During the flight phase of motion, neither the contact nor rolling constraints are active, and thus the corresponding constraint forces can be set to zero, 
$\lambda_1 = \lambda_2 = 0$, to give the following equations of motion.
\begin{subequations}\label{eq:flight_eom}
    \begin{align}
        &I_o\ddot{\phi} = - \tau
        \\[1ex]
        &m_pl_p^2\ddot{\theta} + m_pl_p\left(-\ddot{x}\cos\theta + \ddot{y}\sin\theta + g\sin\theta\right) = \tau
        \\[1ex]
        &m\ddot{x} + m_pl_p\left(\dot{\theta}^2\sin\theta -\ddot{\theta}\cos\theta\right) = 0
        \\[1ex]
        &m\ddot{y} + m_pl_p\left(\dot{\theta}^2\cos\theta + \ddot{\theta}\sin\theta\right) + mg = 0
    \end{align}
\end{subequations}
The center of mass of the system will undergo projectile motion during the flight phase, and thus will follow a parabolic trajectory, but this full system of equations is needed to describe the internal rotations of the system and the trajectory of the geometric center of the hoop. 

\section{Control strategy and numerical simulation}\label{sec:simulation}
\begin{figure}[t]
    \centering
    \begin{subfigure}[b]{\linewidth}
        \includegraphics[width=\linewidth]{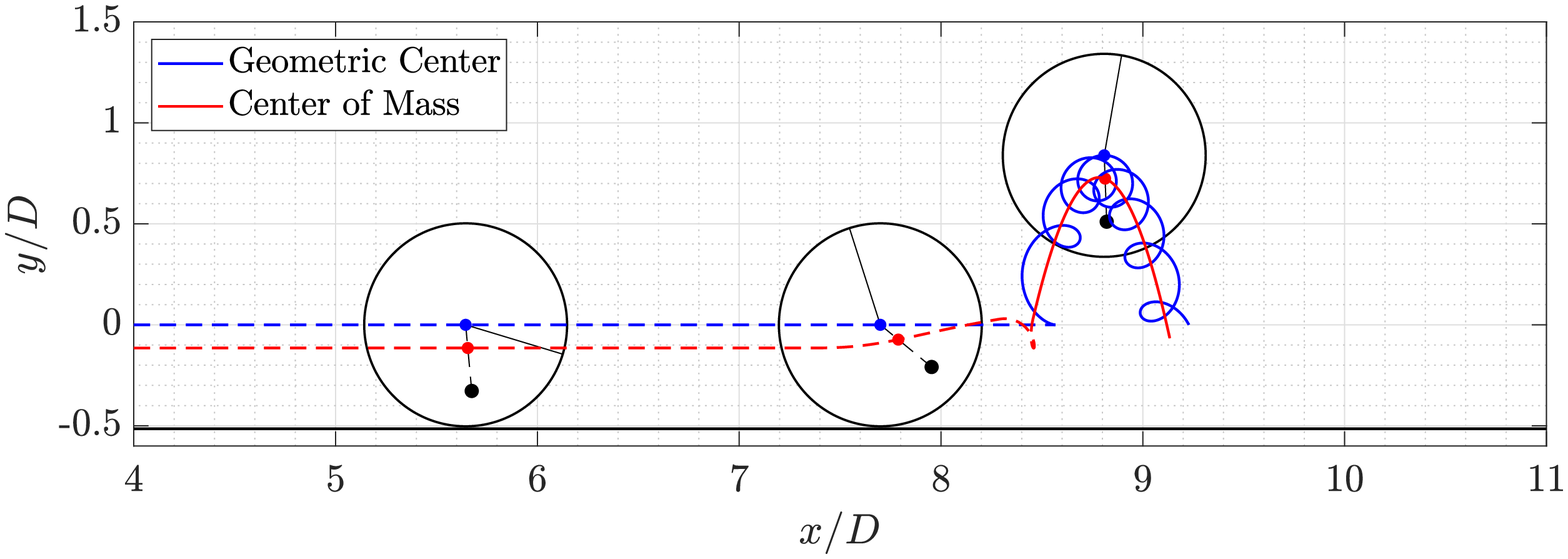}
        \caption{}
        \label{fig:xytrajsim}
    \end{subfigure}
    \begin{subfigure}[b]{0.49\linewidth}
        \includegraphics[width = \linewidth]{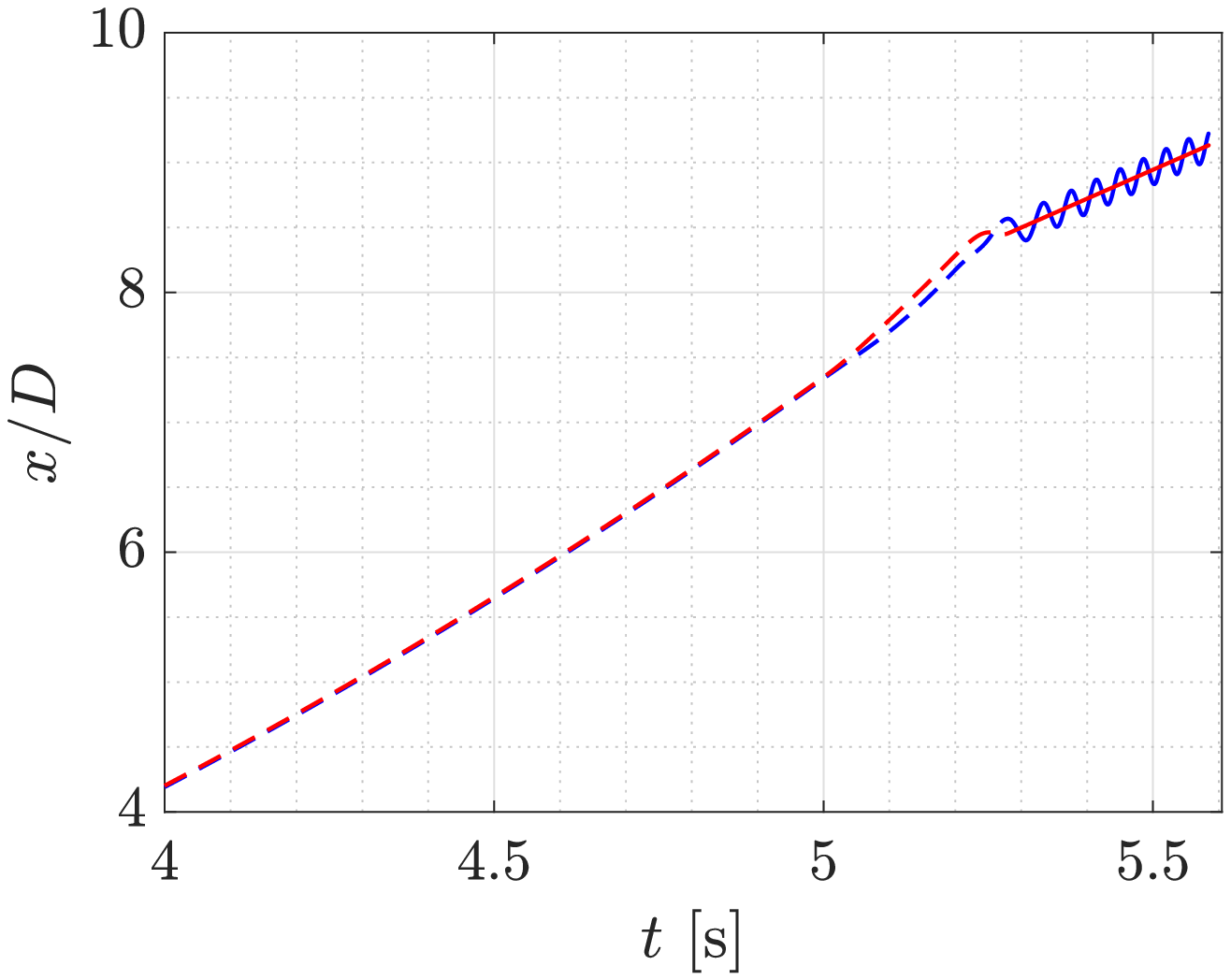}
        \caption{}
        \label{fig:xtrajsim}
    \end{subfigure}
    \begin{subfigure}[b]{0.49\linewidth}
        \includegraphics[width =\linewidth]{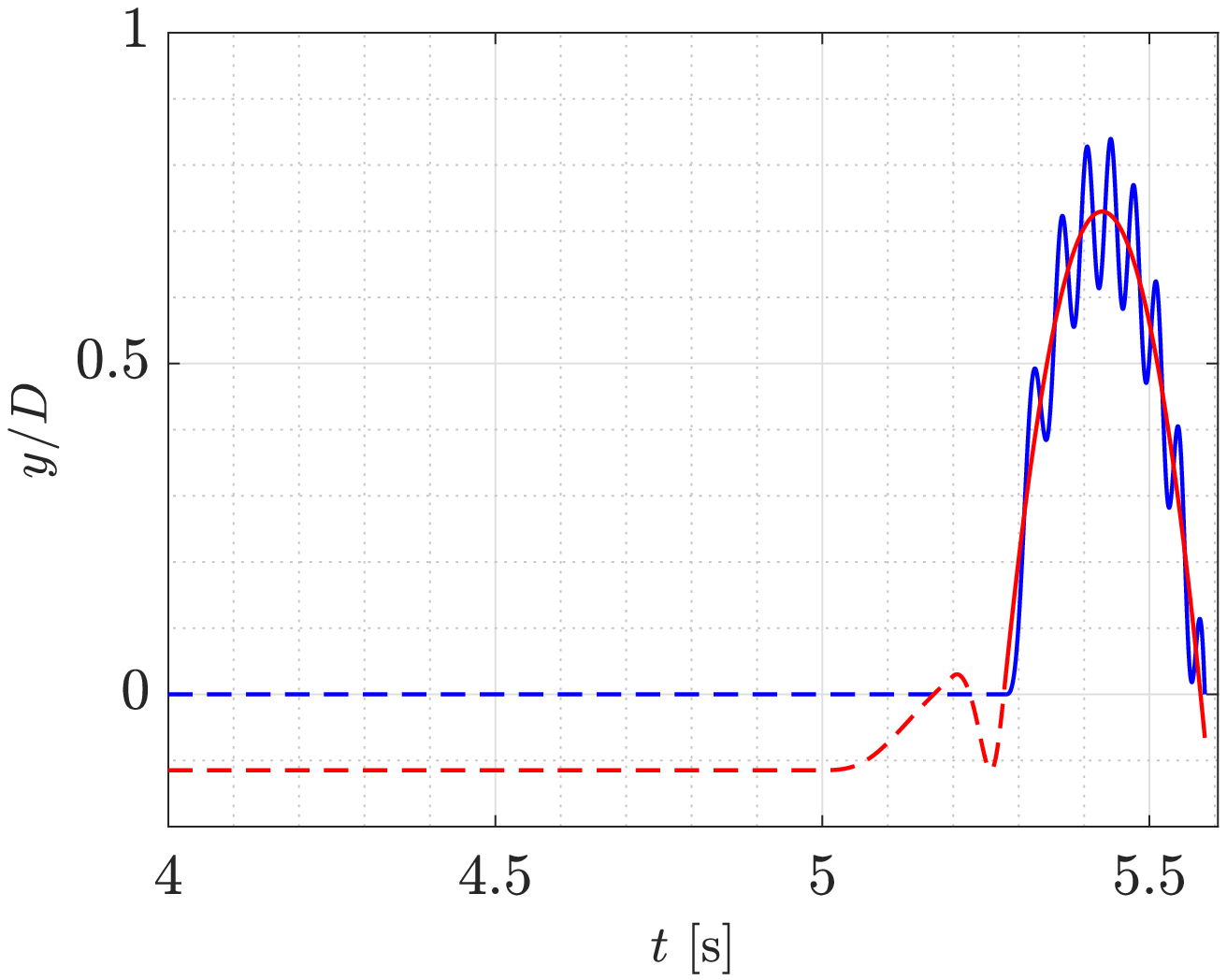}
        \caption{}
        \label{fig:ytrajsim}
    \end{subfigure}
    \begin{subfigure}[b]{0.49\linewidth}
        \includegraphics[width = \linewidth]{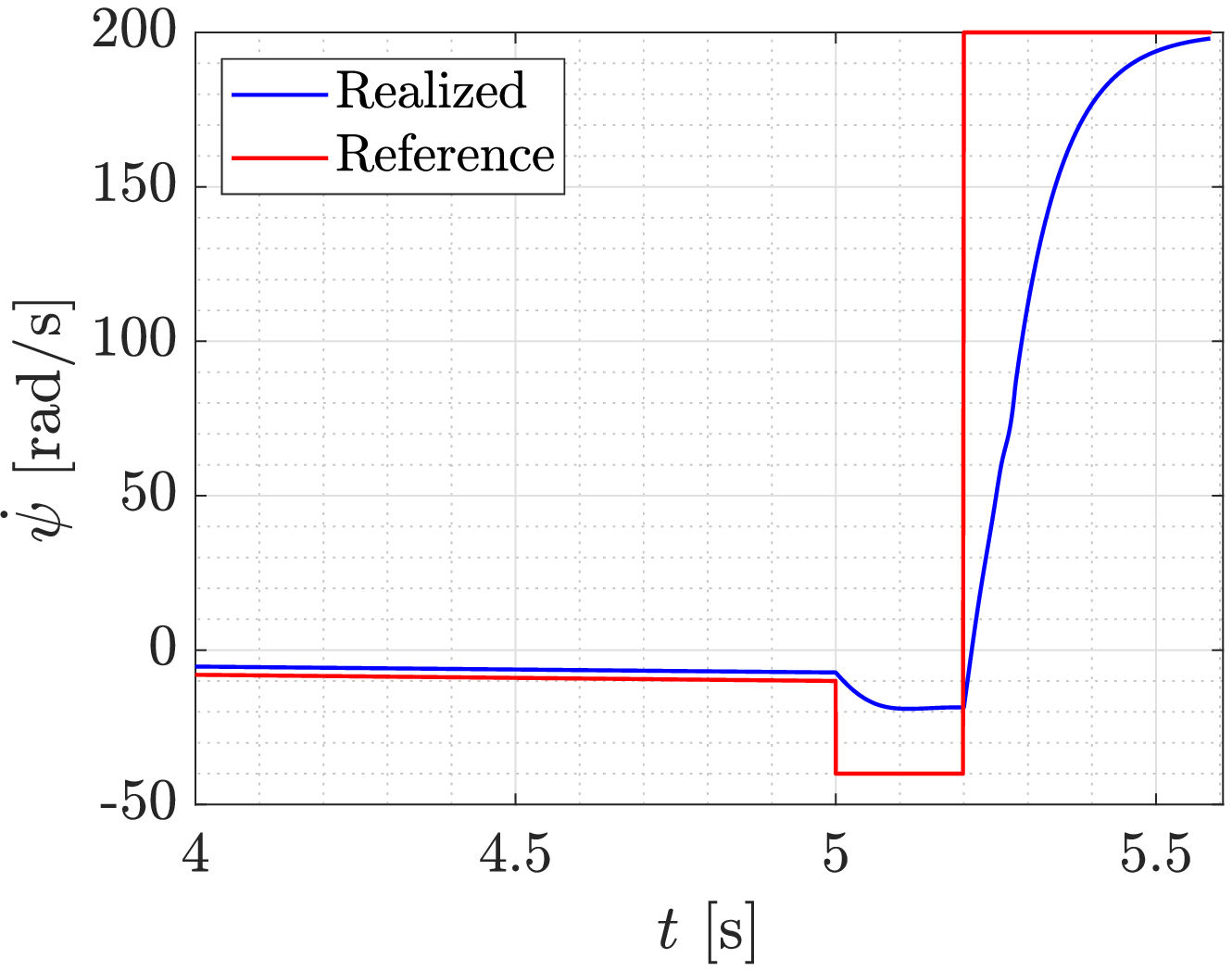}
        \caption{}
        \label{fig:dpsitrajsim}
    \end{subfigure}
    \begin{subfigure}[b]{0.49\linewidth}
        \includegraphics[width = \linewidth]{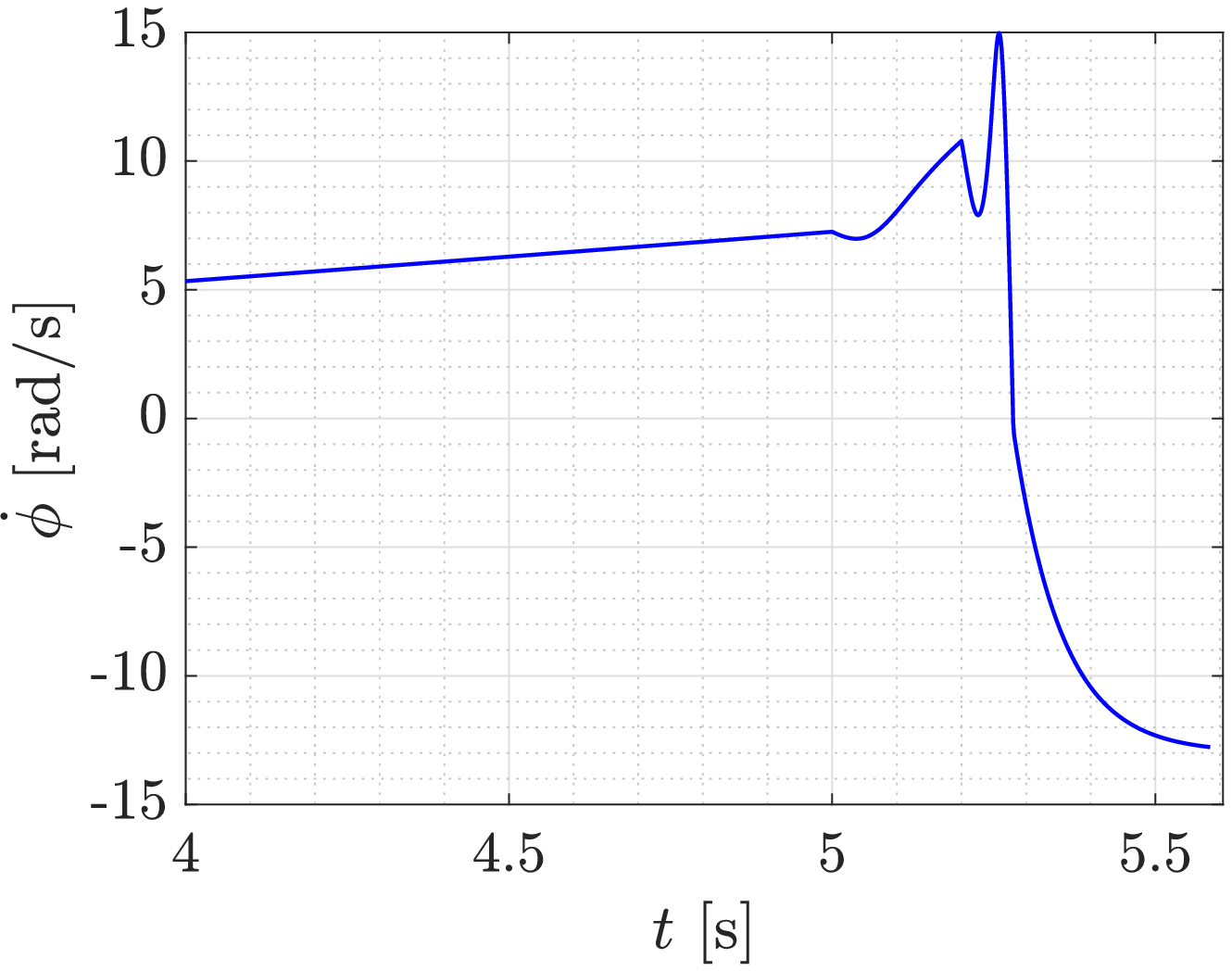}
        \caption{}
        \label{fig:dphitrajsim}
    \end{subfigure}    
    \begin{subfigure}[b]{0.53\linewidth}
        \includegraphics[width = \linewidth]{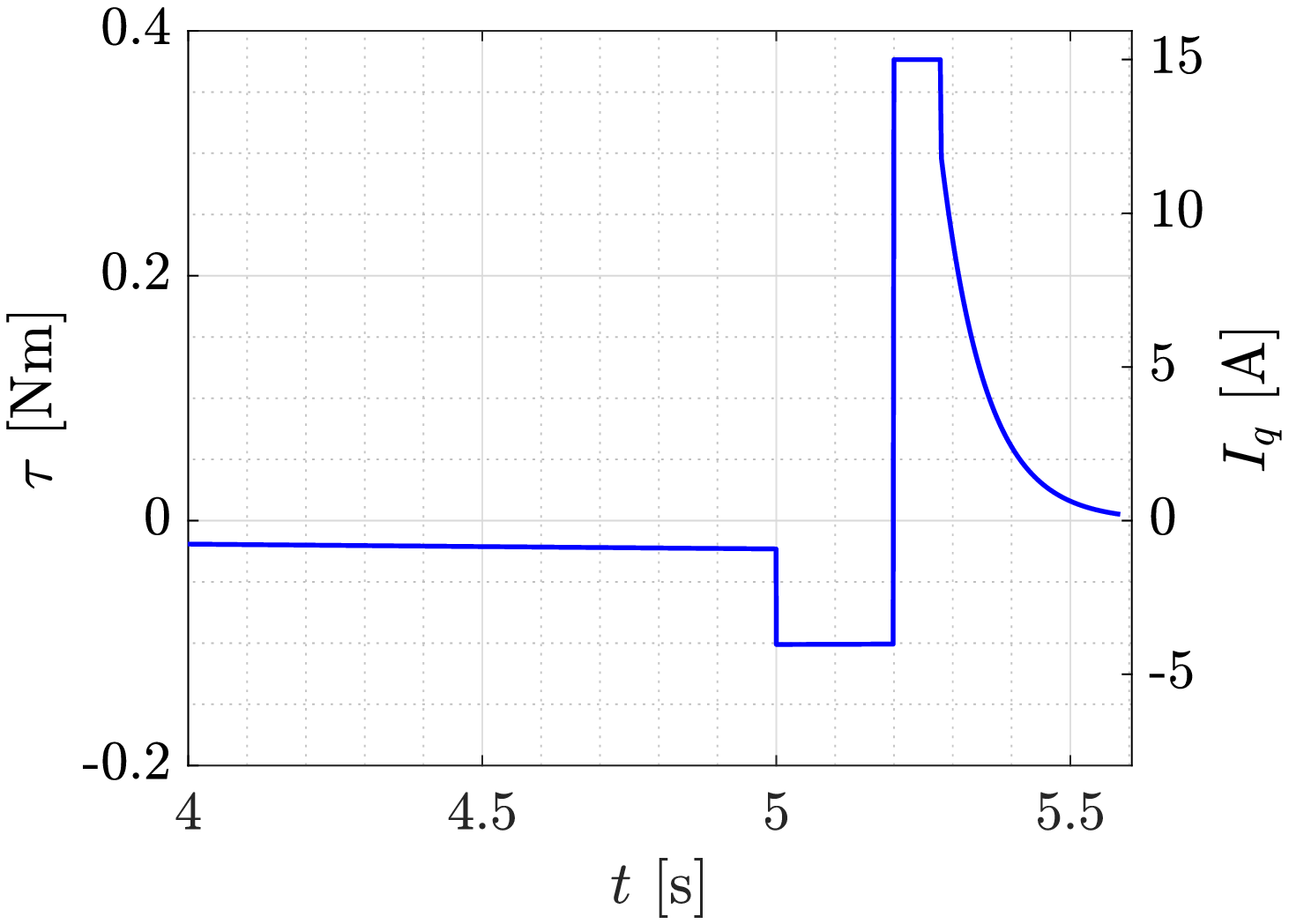}
        \caption{}
        \label{fig:tautrajsim}
    \end{subfigure}
    \begin{subfigure}[b]{0.45\linewidth}
        \begin{tikzpicture}
            \node[anchor=south west,inner sep=0] (f4) at (0,0) {\includegraphics[width = \linewidth]{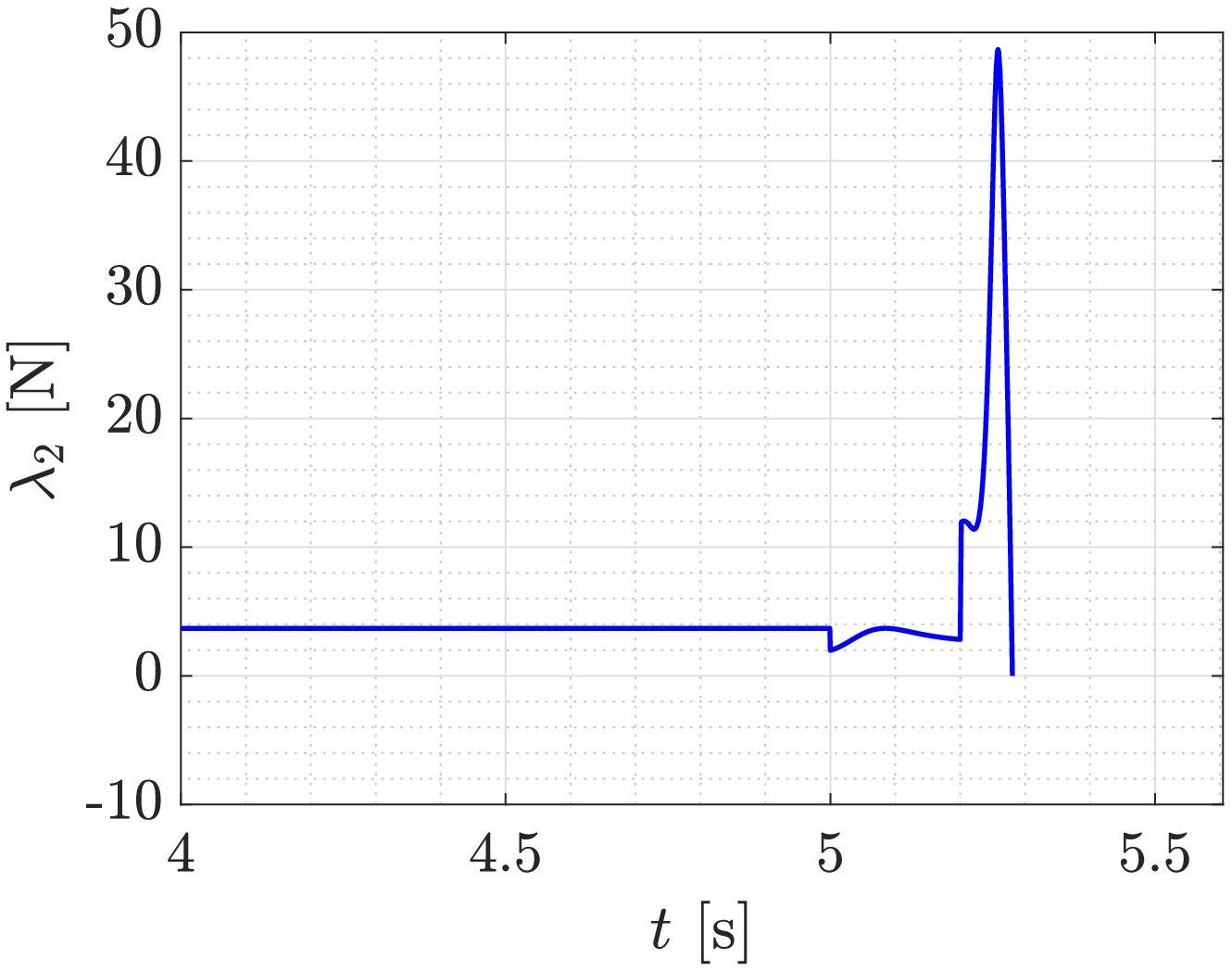}};
            \node[scale = 0.6,anchor=north,text width = 2cm] (T) at (1.5,2.5) {$\lambda_2 = 0$\\ Jump occurs};
            \draw[-latex] (T)--(3.2,0.95);
            \draw (3.21,0.94) circle [radius=1.5pt];
        \end{tikzpicture}
        \caption{}
        \label{fig:Ntrajsim}
    \end{subfigure}
    \caption{Numerical simulation of the jumping wheel system. Dashed lines indicate the rolling phase while solid lines indicate the flight phase. (a) Spatial trajectory of the system center of mass and geometric center. (b)-(c) trajectories of these points over time. (d) relative angular velocity between the pendulum and the wheel (e) angular velocity of the wheel(f) torque values $\tau$ and corresponding motor currents $I_q$. (g) normal reaction $\lambda_2$ during rolling. } 
    \label{fig:NumSim}
    \vspace{-1em}
\end{figure}

To numerically simulate this system, we implement the equations of motion described in Sec. \ref{sec:model} in \texttt{MATLAB}.  To demonstrate the rolling, jumping motion, we consider the case where the robot begins at rest with the pendulum in the vertically downward position and the robot will be controlled using a proportional controller to choose the torque $\tau$ based on the error from a reference signal for the angular velocity of the pendulum relative to the wheel.  
We denote this angular velocity as $\dot{\psi} = \dot{\theta} - \dot{\phi}$.  Such a simulation is shown in Fig. \ref{fig:NumSim} and is explained in more detail in Sec. \ref{sec:numsim}, but first we will discuss the control strategies implemented to achieve rolling and jumping motions. 
\subsection{Rolling strategy}
To achieve a smooth rolling motion from rest, a torque is applied to lift the pendulum in the desired direction of motion.  The resulting offset of the pendulum will cause the wheel to roll to restore the pendulum to the stable, downward equilibrium, thus leading the wheel to accelerate in the direction of the pendulum. 
Once the wheel is in rolling motion, it's velocity can be maintained by driving the pendulum so that it remains in the downward position.  This is done by driving the pendulum so that its angular velocity relative to the wheel offsets the velocity of the wheel; that is, $\dot{\psi} = -\dot{\phi}$ so that $\dot{\theta} = 0$. This has the effect of holding the pendulum in the downward position in the spatial frame, but requires continuous actuation to move the pendulum relative to the rolling wheel.
Following a similar reasoning, the wheel speed can be increased in a given direction by driving the pendulum relative to the wheel in such a way that the pendulum is displaced in that direction in the spatially fixed frame. 

\subsection{Jumping strategy}
As mentioned previously in Sec. \ref{sec:model}, jumping motion will tend to occur when the pendulum has large angular velocity (relative to the spatially fixed frame) and the pendulum nears the upward configuration ($\theta \approx n\pi$ and $\dot{\theta}^2 \gg 1)$, which causes the $m_pl_p\dot{\theta}^2\cos\theta$ term in Eq. \ref{eq:normal_reaction} to become large, which leads the normal reaction $\lambda_2$ to vanish with the center of mass velocity having a positive vertical component. Therefore, to achieve a jump, the pendulum should be driven at high angular velocity toward the upward configuration once the desired rolling velocity is reached. 

Increasing the horizontal distance traversed during the jump requires increasing the horizontal component of the velocity of the center of mass at the instant that the loss of contact occurs, as from that point forward, the center of mass behaves as a projectile, and thus its displacement is a function of its initial position and velocity alone.  
The horizontal component of the center of mass velocity during rolling is $\dot{\phi}R - \frac{m_pl_p}{m_o+m_p}\dot{\theta}\cos\theta\,.$
Thus, the horizontal displacement can be increased by 
increasing the rolling speed of the wheel, $\dot{\phi}R$.  
However, 
at higher rolling speed there is less time for the pendulum to accelerate from its downward position during rolling to the upward configuration necessary for a jump. 
Therefore, better jumping can be achieved by first initiating ``swing-up'' phase in which the pendulum is brought to the upward position at a low angular velocity 
before actuating the pendulum with maximum torque in the opposite direction so that the pendulum has 
more time to accelerate and maximize $\dot{\theta}^2$.

\subsection{Numerical simulation}\label{sec:numsim}
As proof of concept, we implement a numerical simulation of the proposed control strategy.  In this simulation, the wheel is started from rest with the pendulum in the vertically downward position. The torque is chosen by applying a proportional controller to track a reference relative angular velocity, $\dot{\psi}_{\text{ref}}$.  This reference angular velocity is chosen to give a 5s rolling period in which the wheel accelerates from rest to an angular velocity of approximately 10 rad/s.
For this, the reference angular velocity is chosen to be linear in time.  
Following this rolling phase, the swing-up phase is initiated.
Finally, the pendulum is actuated clockwise at a large angular velocity in order to achieve a jump as the pendulum nears the upward configuration. 
Specifically, the relative angular velocity reference command is chosen as follows:
\begin{equation}\label{eq:vert_jump_dpsi}
\dot{\psi}_\text{ref}(t) =
\begin{cases}
-2t & t < 5\\
-40 & 5.0 < t < 5.2 \\
200 & t > 5.2
\end{cases}
\end{equation}
where the times are given in seconds and the angular velocities are in rad/s.
All parameters used in this simulation are selected to be representative of the robot described in Sec \ref{sec:robot_design}.

The simulation is conducted by first simulating the rolling dynamics given by Eqs. \ref{eq:rolling_eom} until the normal reaction vanishes with a positive vertical component of the center of mass velocity.
This state is then taken as the initial condition for the flight phase and the system is simulated from there using the flight dynamics given by Eqs. \ref{eq:flight_eom}.
The results of this simulation are shown in Fig. \ref{fig:NumSim}. 
It can be seen from Figs. \ref{fig:xytrajsim}-\ref{fig:ytrajsim} that while in the air, the center of the hoop follows a cycloidal trajectory, revolving about the trajectory of the center of mass, which follows a parabolic trajectory.  
The simulation is ended when the $y$ coordinate of the center of the wheel returns to zero, indicating that the wheel has landed on the ground.  
Further simulation beyond this point would require additional modelling of the collision dynamics associated with the impact with the ground.

	\begin{figure}
		    \centering
      \begin{subfigure}[b]{\linewidth} 
      \centering
      \begin{tikzpicture}
        \node[anchor=south west,inner sep = 0] (image) at (0,0) {
            \includegraphics[width =0.8\linewidth,trim={7.5cm 6.5cm 11cm 0},clip]{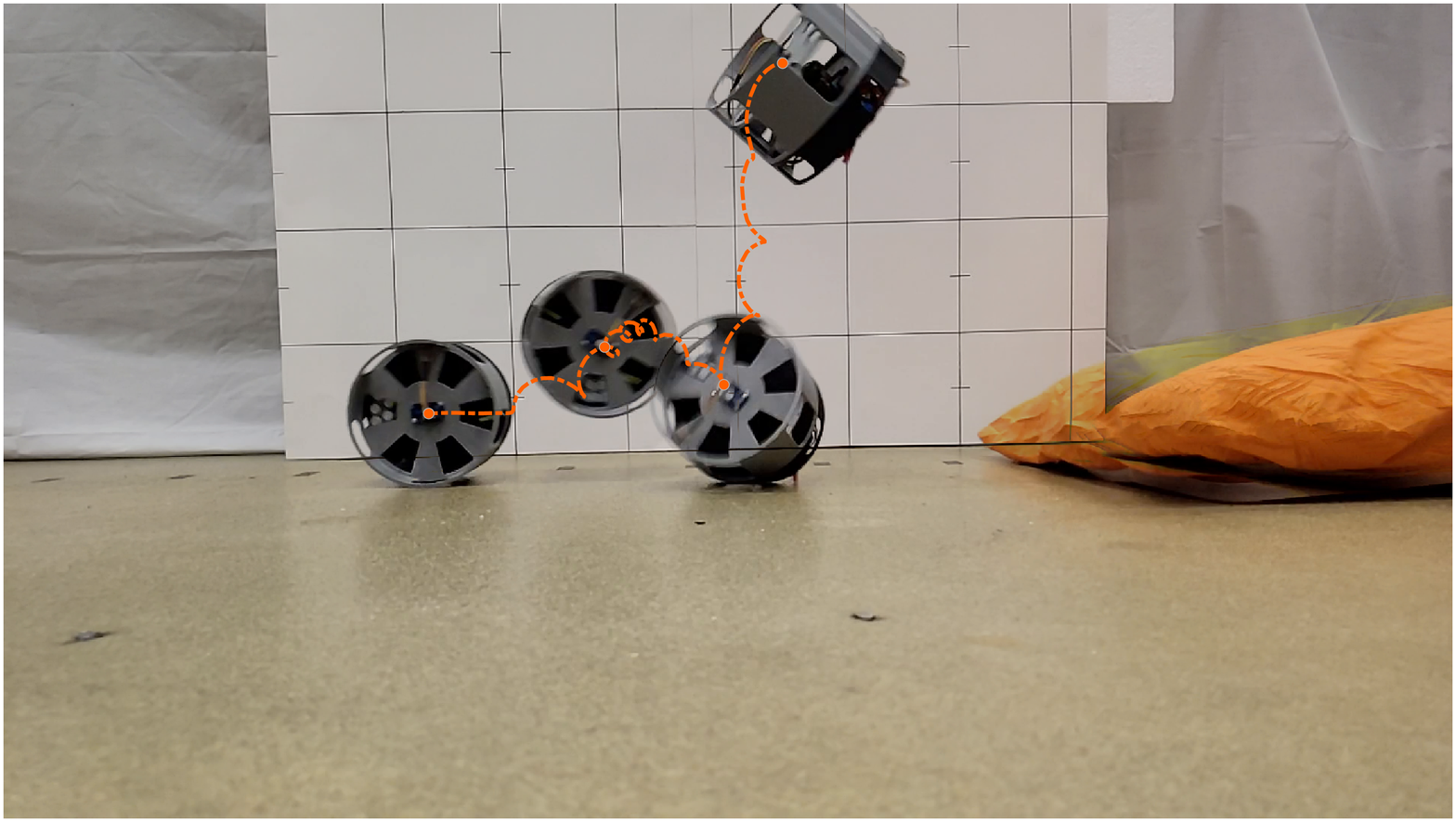}};
        \draw (image.north east) rectangle (image.south west);
        \draw[stealth-stealth, very thick] (6.5,1.65) -- (6.5,5.1);
        \draw (5.5,1.65) -- ++ (1.2,0.0);
        \draw (5.5,5.1) -- ++ (1.2,0.0);
        \node at (5.3,3.2)[right,black,text width = 1.0cm,align=right]{2.4 body lengths};
      \end{tikzpicture}\\
      \caption{}\vspace{-1ex}
      \label{fig:vert_jump_traj_pic}
      \end{subfigure}
      \\[1em]
      \begin{subfigure}[b]{0.49\linewidth} 
      \includegraphics[width = \linewidth]{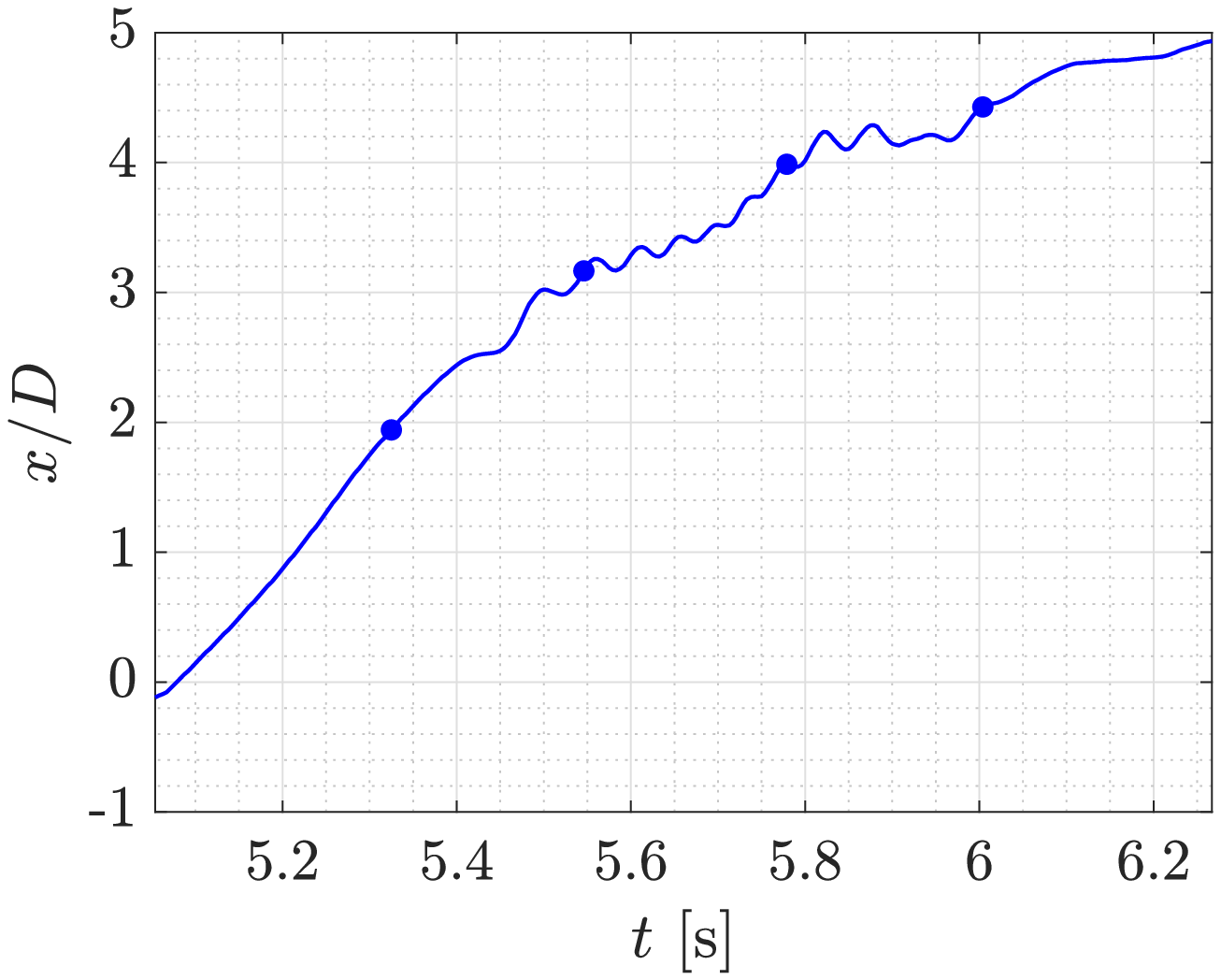}
      \caption{}\label{fig:vert_jump_xtraj}
      \end{subfigure}
      \begin{subfigure}[b]{0.49\linewidth} 
      \includegraphics[width = \linewidth]{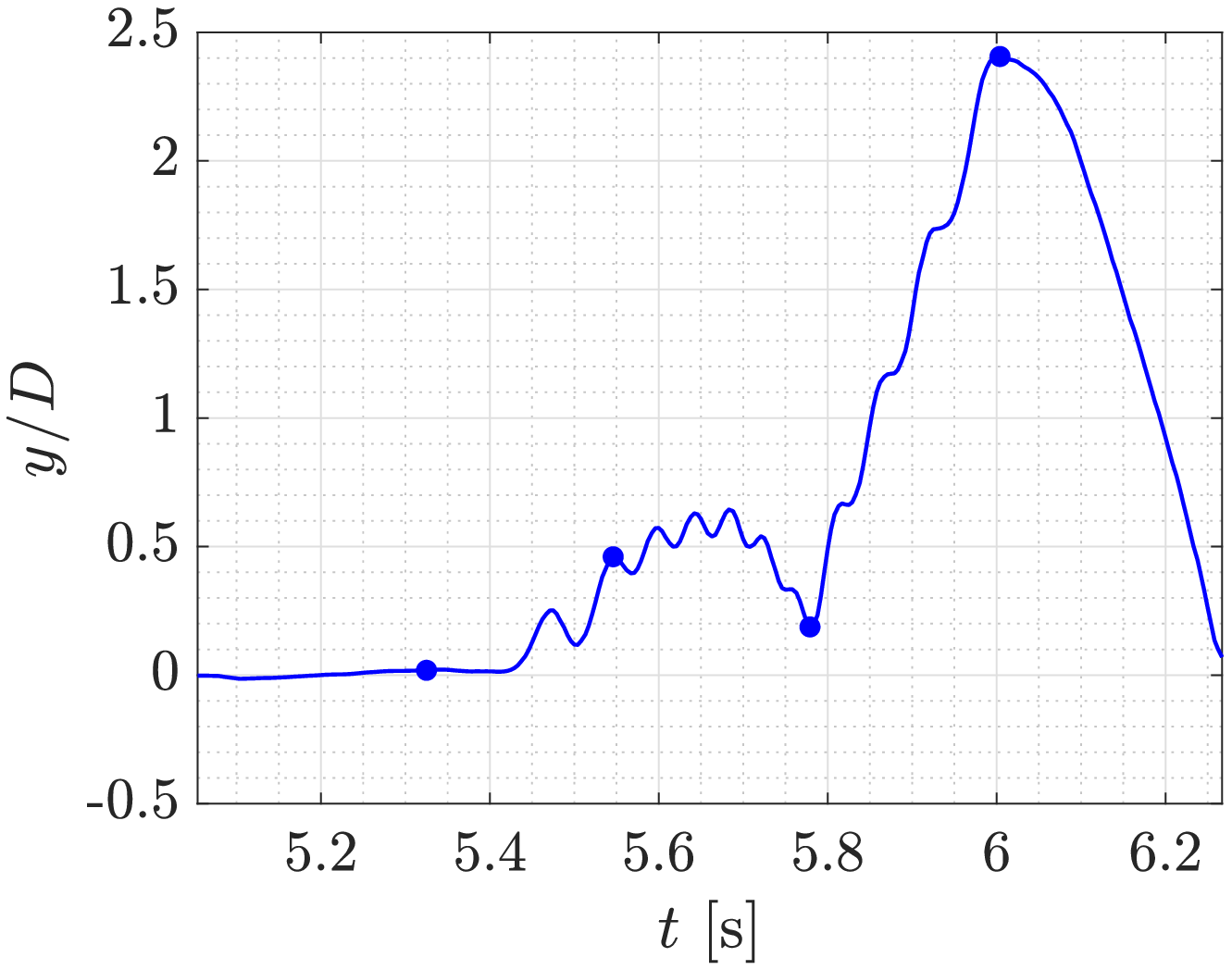}
      \caption{}\label{fig:vert_jump_ytraj}
      \end{subfigure}
      \begin{subfigure}[b]{0.49\linewidth}
        \includegraphics[width = \linewidth]{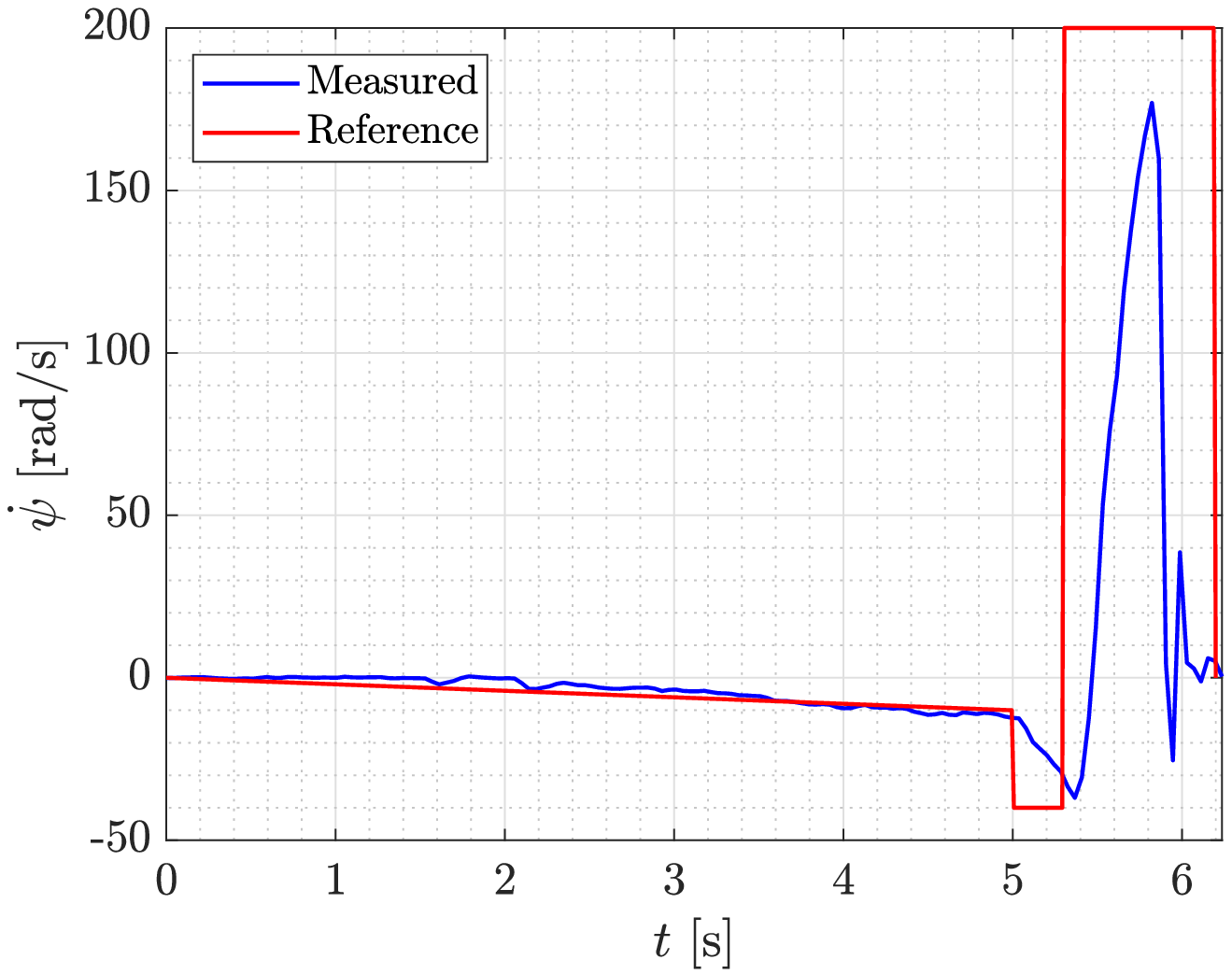}
        \caption{}
        \label{fig:vert_dpsi}
    \end{subfigure}
        \begin{subfigure}[b]{0.49\linewidth}
            \includegraphics[width = \linewidth]{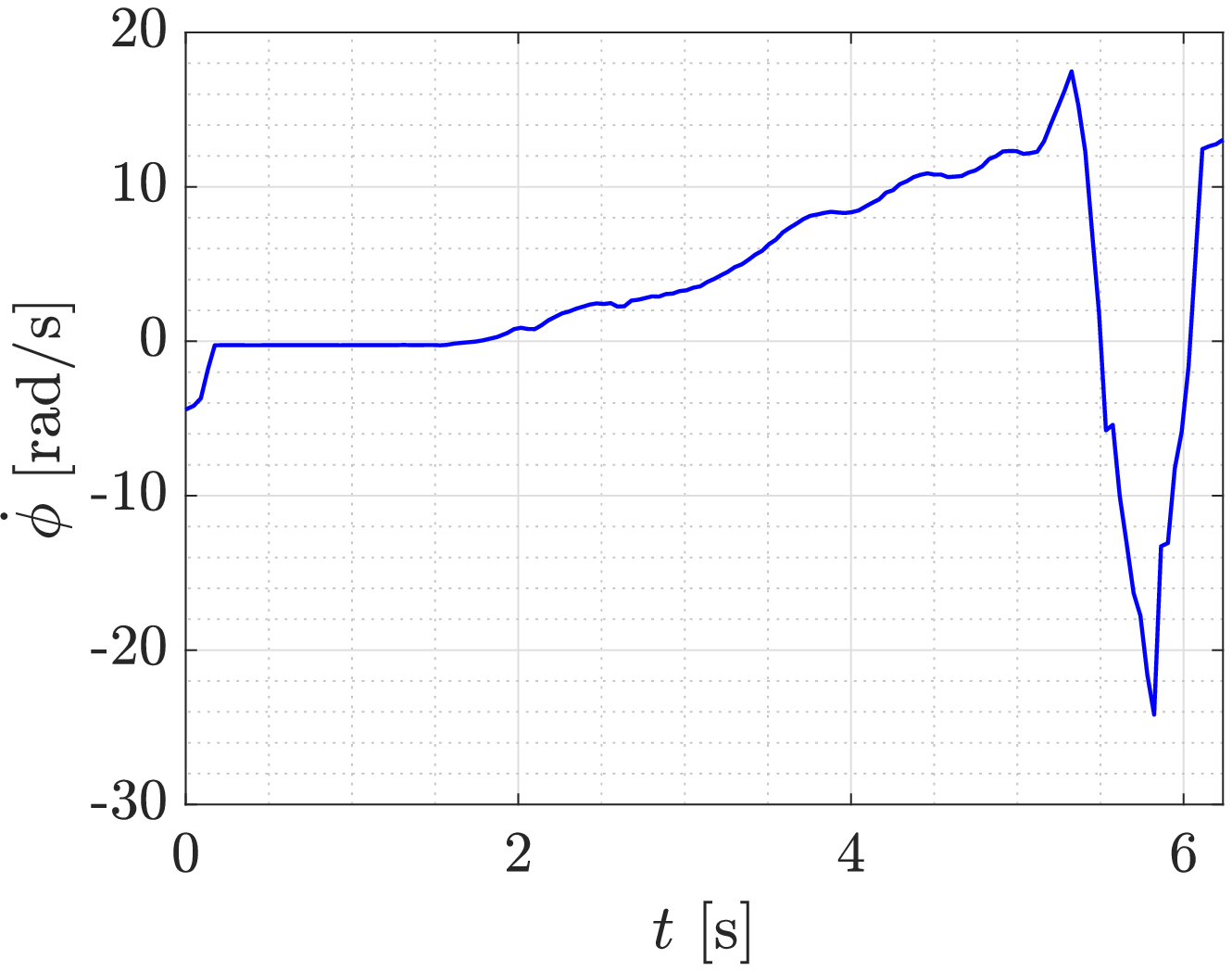}
            \caption{}
            \label{fig:vert_dphi}
        \end{subfigure}
		    \caption{Experimental results for vertical jumping. (a) Four still images from a video of the jumping motion overlaid to illustrate the robot trajectory. (b)-(c) Trajectory of a point on the front of the wheel over time. (d) Relative angular velocity between pendulum and wheel. (e) Angular velocity of the wheel.}
            \vspace{-1ex}
		    \label{fig:vert_jump_exp}
		\end{figure}
\begin{figure*}
      \centering 
            \begin{tikzpicture}%
            \node[anchor=south west,inner sep=0] (f) at (0,0) {\includegraphics[width = 0.19\linewidth]{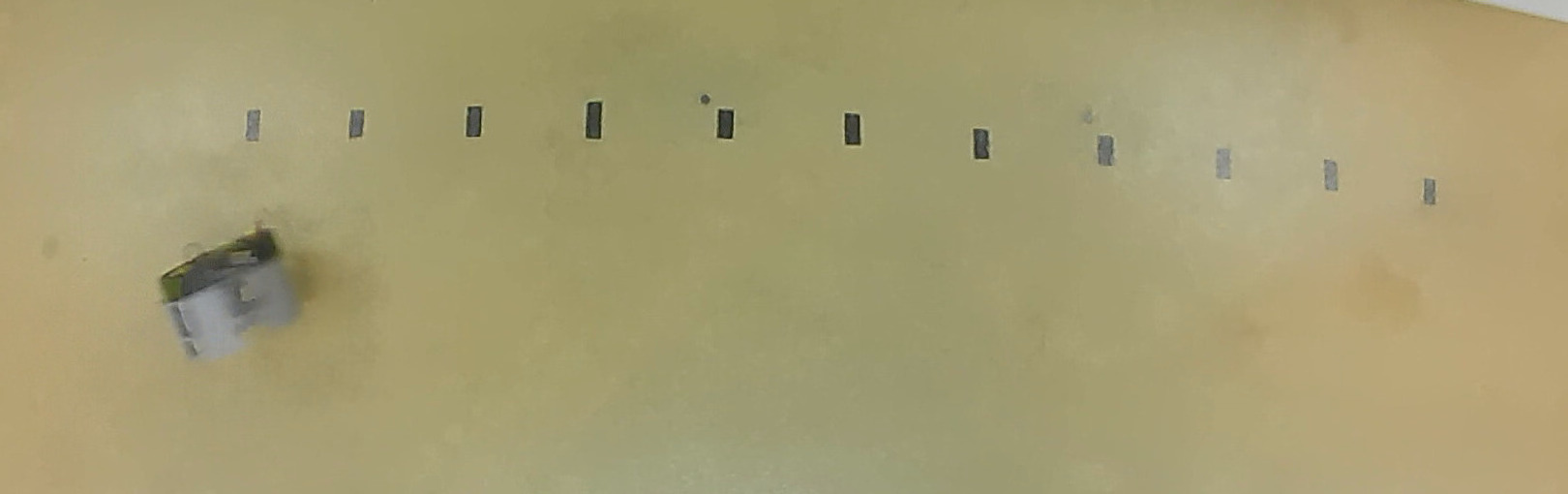}};
            \draw [very thin] (f.north east) rectangle (f.south west);
            \draw [thin, red] (0.49,0.9) -- ++(0,-0.25); 
        \end{tikzpicture}
        \begin{tikzpicture}%
            \node[anchor=south west,inner sep=0] (f) at (0,0) {\includegraphics[width = 0.19\linewidth]{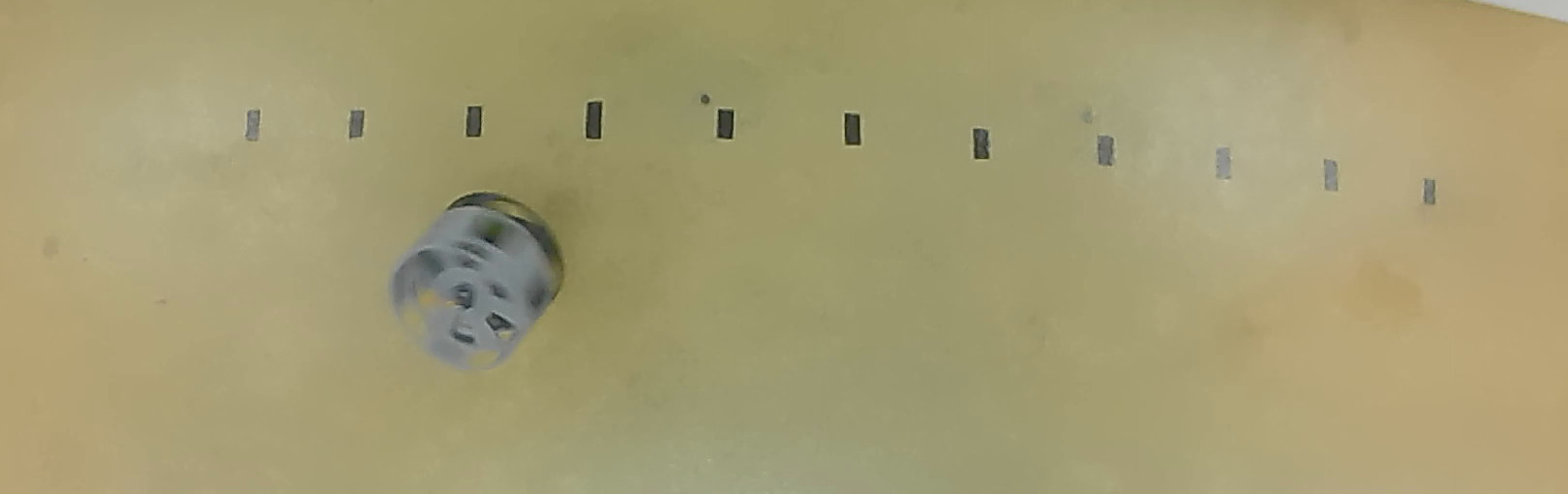}};
            \draw [very thin] (f.north east) rectangle (f.south west);
        \end{tikzpicture}
        \begin{tikzpicture}%
            \node[anchor=south west,inner sep=0] (f) at (0,0) {\includegraphics[width = 0.19\linewidth]{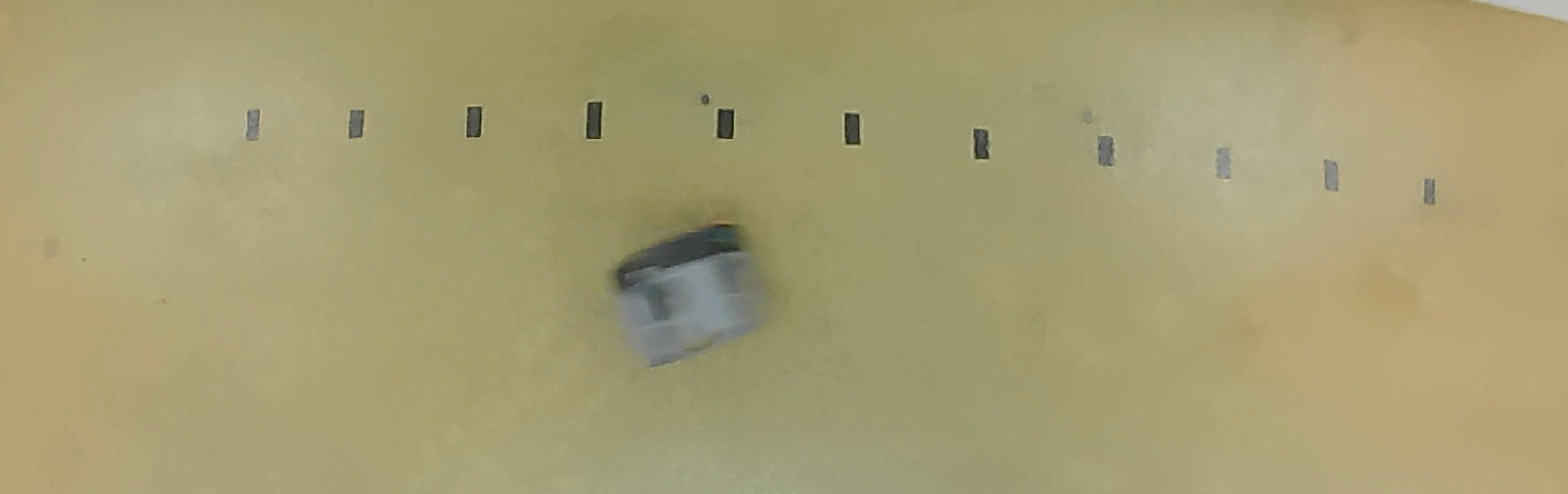}};
            \draw [very thin] (f.north east) rectangle (f.south west);
            \draw [thin, red] (0.49,0.9) -- ++(0,-0.25); 
            \draw [thin, red] (1.5,0.9) -- ++(0,-0.25); 
            \draw [thin, red, latex-latex] (0.49,0.9-0.125) -- (1.5,0.9-0.125);
        \end{tikzpicture}
        \begin{tikzpicture}%
            \node[anchor=south west,inner sep=0] (f) at (0,0) {\includegraphics[width = 0.19\linewidth]{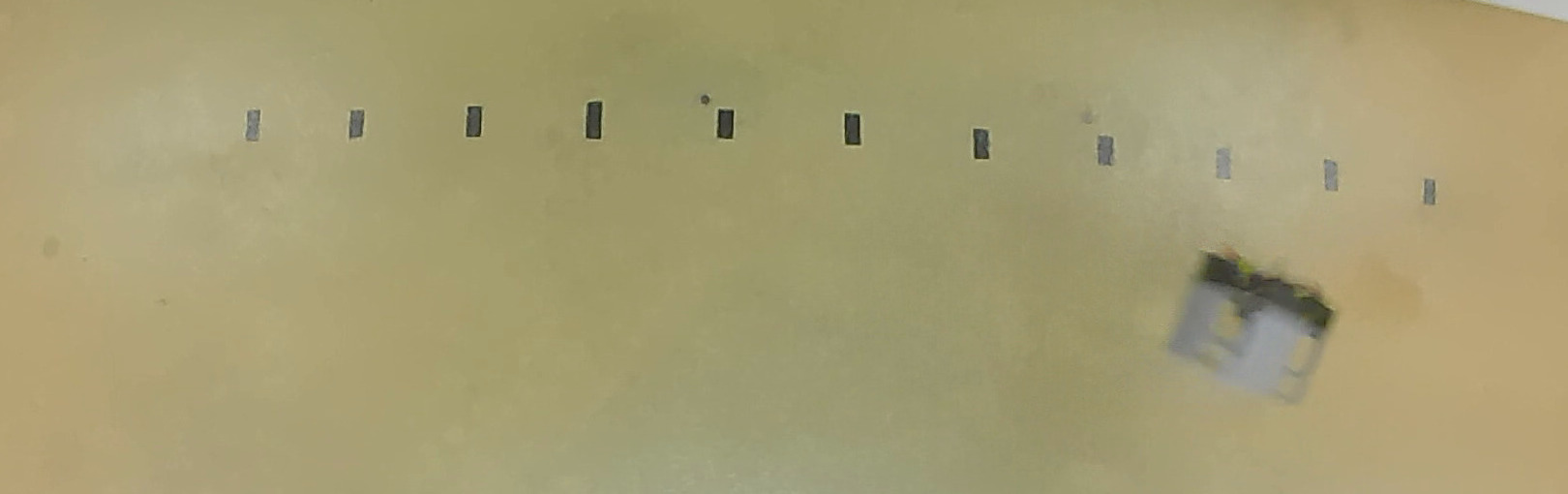}};
            \draw [very thin] (f.north east) rectangle (f.south west);
        \end{tikzpicture}
        \begin{tikzpicture}%
            \node[anchor=south west,inner sep=0] (f) at (0,0) {\includegraphics[width = 0.19\linewidth]{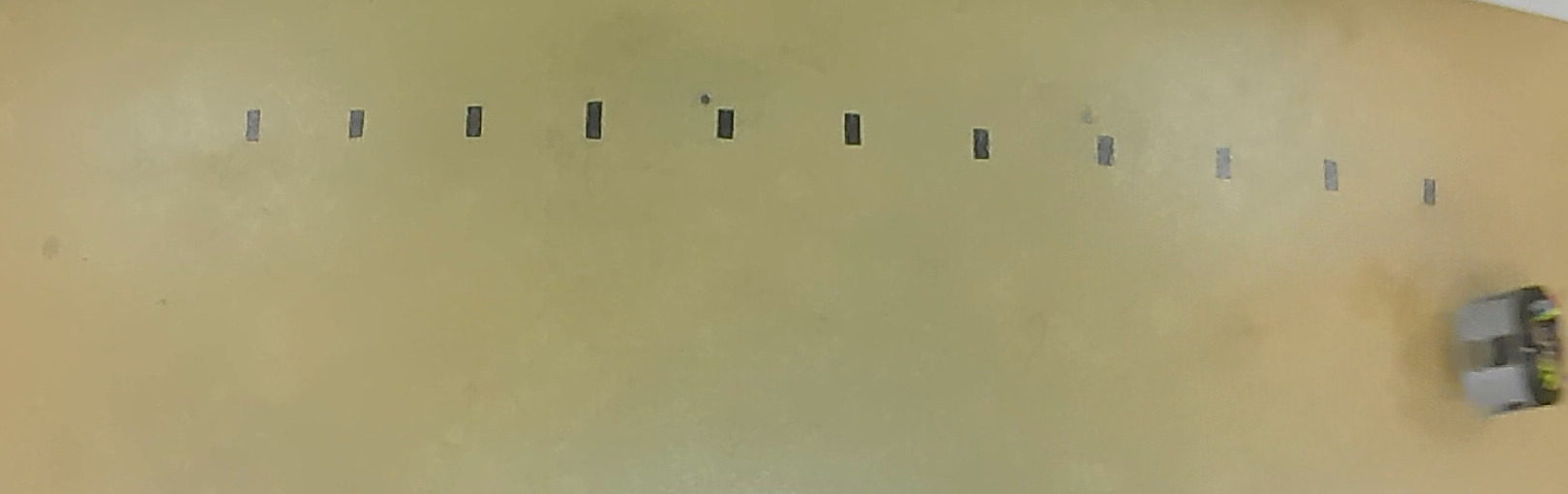}};
            \draw [very thin] (f.north east) rectangle (f.south west);
            \draw [thin, red] (1.5,0.9) -- ++(0,-0.35); 
            \draw [thin, red] (3.25,0.9) -- ++(0,-0.35); 
            \draw [thin, red, latex-latex] (1.5,0.9-0.125) -- (3.25,0.9-0.125);
        \end{tikzpicture}
            \\[1ex]
            \begin{tikzpicture}%
            \node[anchor=south west,inner sep=0] (f) at (0,0) {\includegraphics[width = 0.19\linewidth]{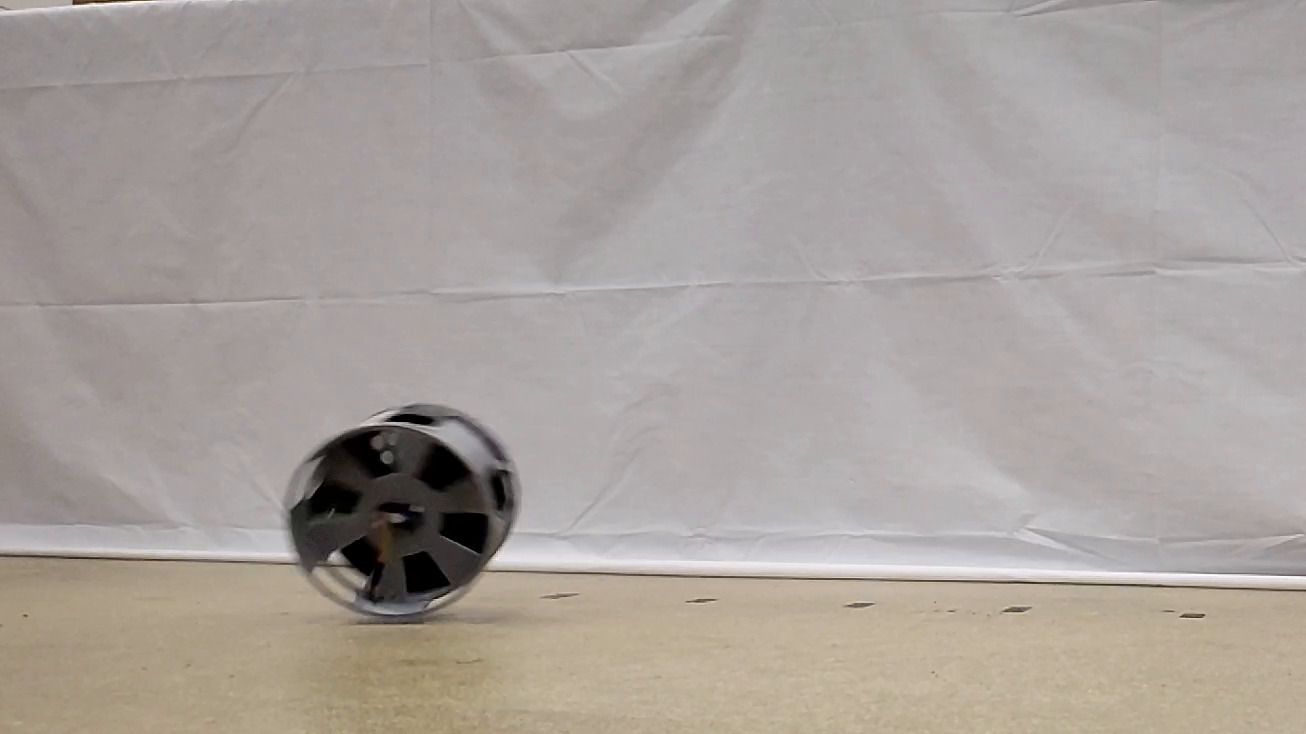}};
                \draw [very thin] (f.north east) rectangle (f.south west);
                \node (C) at ($(f.south east)!0.5!(f.south west)+(0,-0.3)$) {(a)};
            \end{tikzpicture}
            \begin{tikzpicture}%
                \node[anchor=south west,inner sep=0] (f) at (0,0) {\includegraphics[width = 0.19\linewidth]{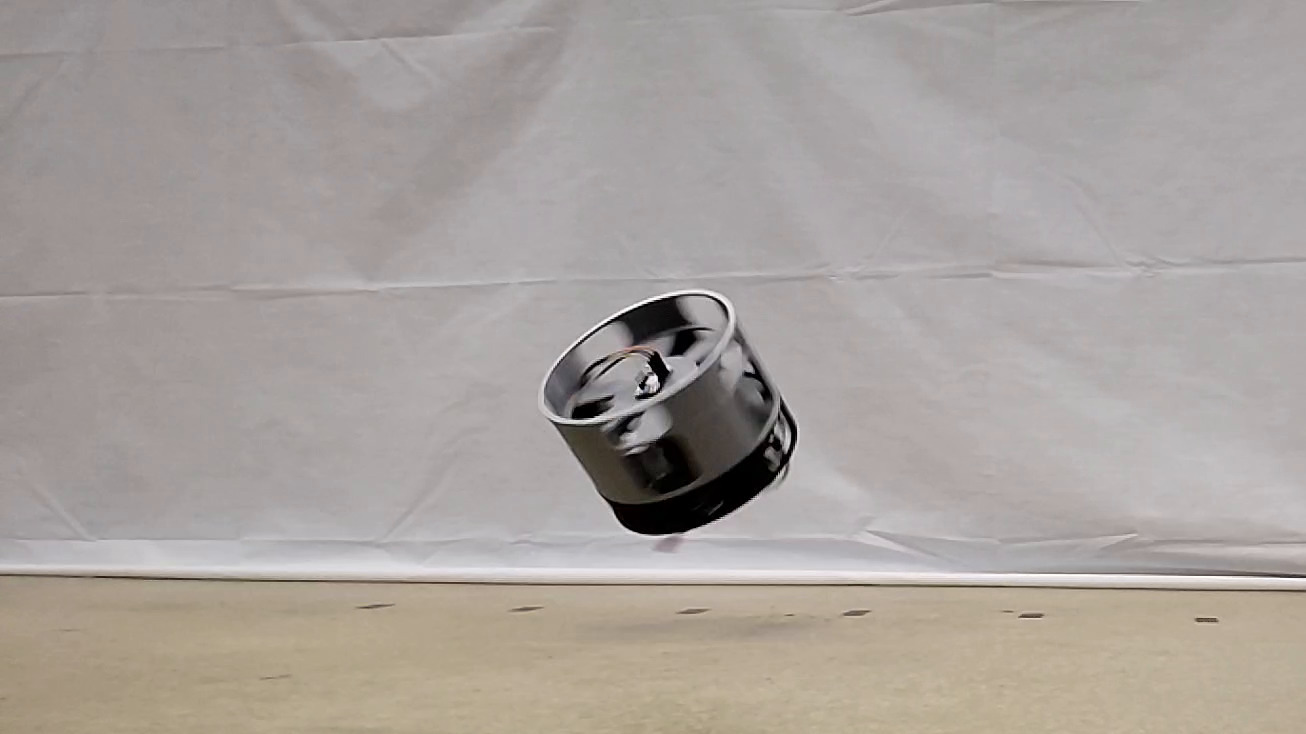}};
                \draw [very thin] (f.north east) rectangle (f.south west);
                \node (C) at ($(f.south east)!0.5!(f.south west)+(0,-0.3)$) {(b)};
                \end{tikzpicture}
            \begin{tikzpicture}%
                \node[anchor=south west,inner sep=0] (f) at (0,0) {\includegraphics[width = 0.19\linewidth]{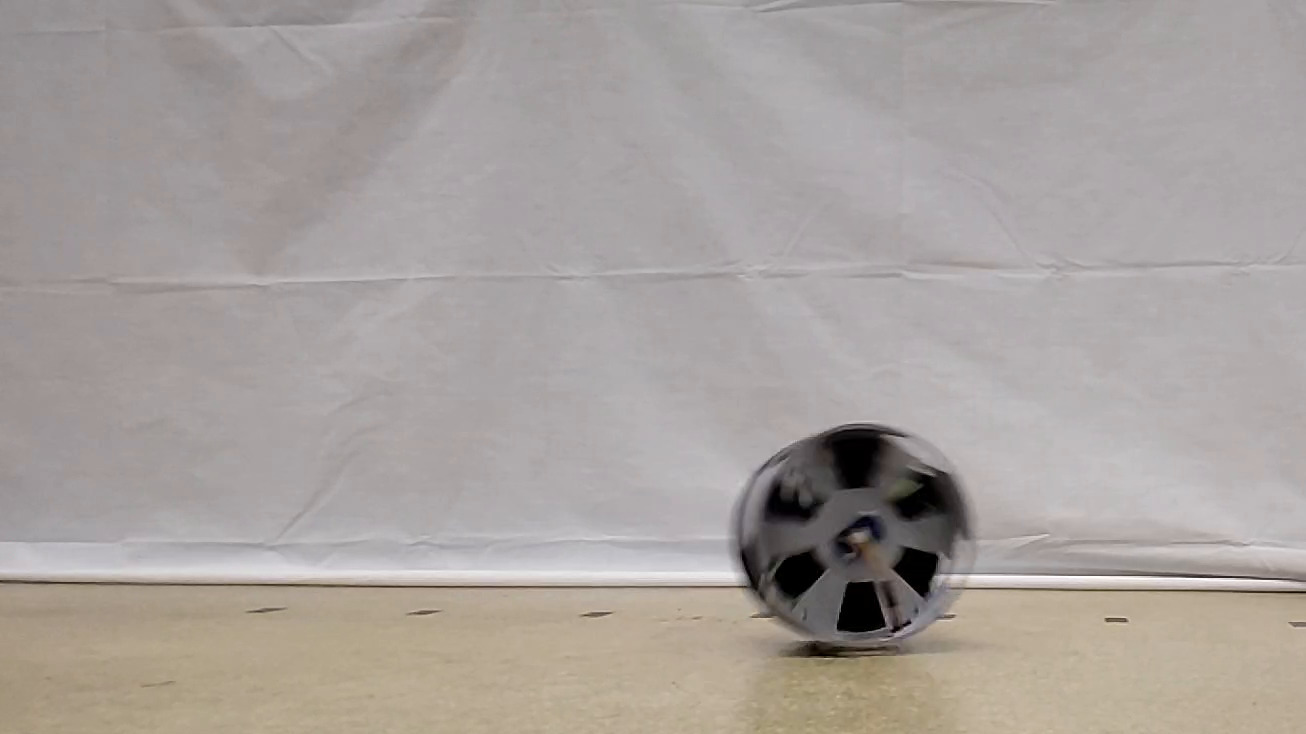}};
                \draw [very thin] (f.north east) rectangle (f.south west);
                \node (C) at ($(f.south east)!0.5!(f.south west)+(0,-0.3)$) {(c)};
            \end{tikzpicture}
            \begin{tikzpicture}%
                \node[anchor=south west,inner sep=0] (f) at (0,0) {\includegraphics[width = 0.19\linewidth]{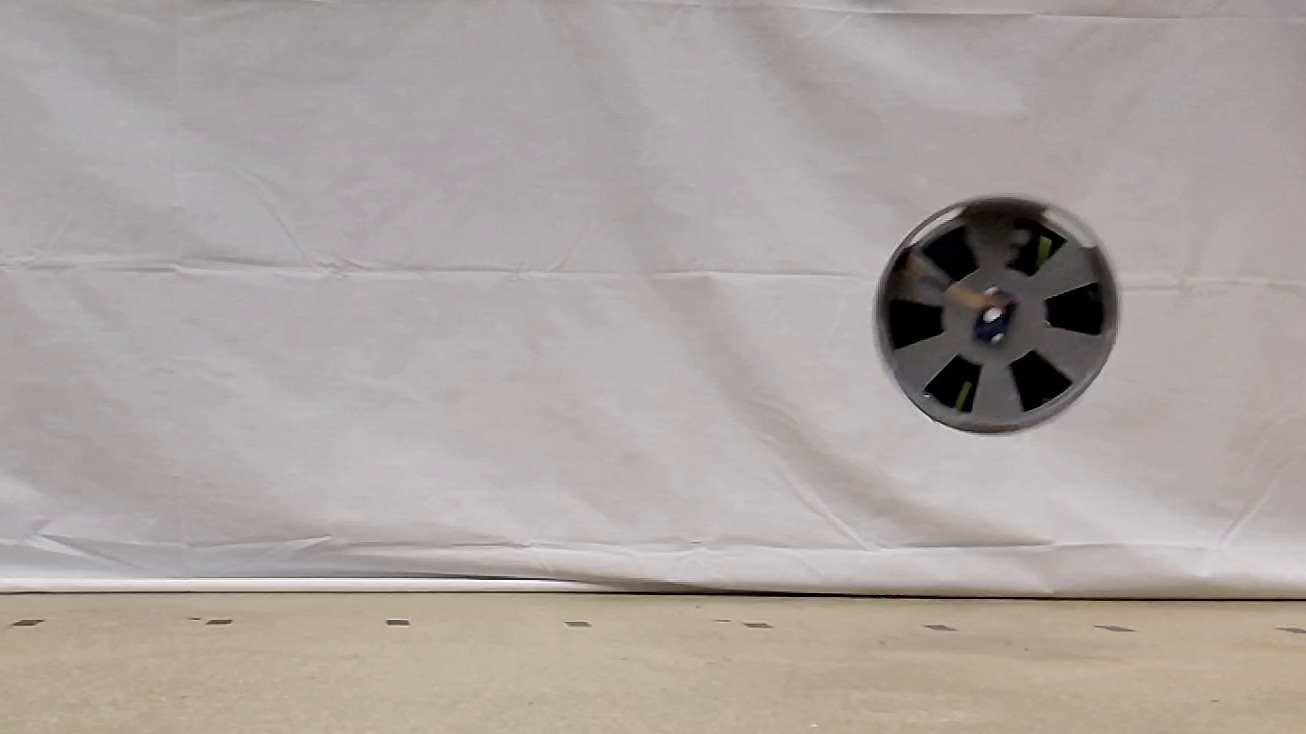}};
                \draw [very thin] (f.north east) rectangle (f.south west);
                \node (C) at ($(f.south east)!0.5!(f.south west)+(0,-0.3)$) {(d)};
            \end{tikzpicture}
            \begin{tikzpicture}%
                \node[anchor=south west,inner sep=0] (f) at (0,0) {\includegraphics[width = 0.19\linewidth]{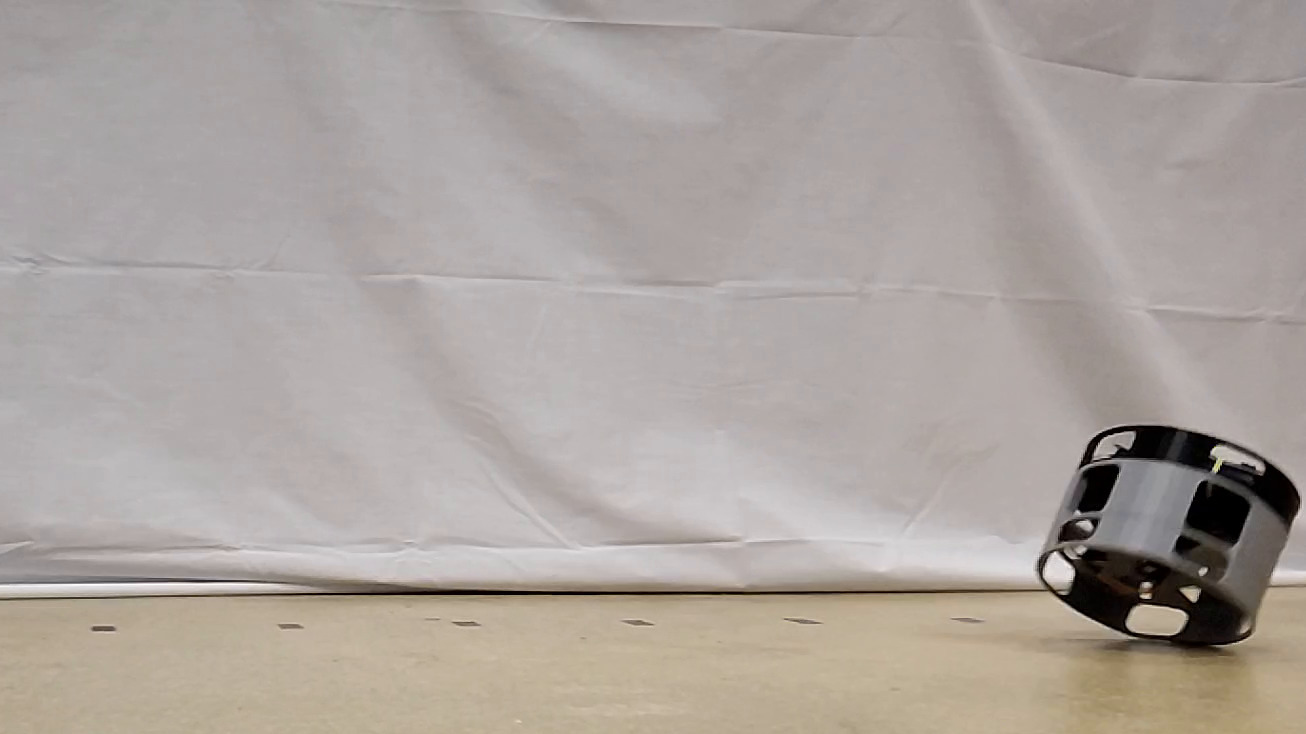}};
                \draw [very thin] (f.north east) rectangle (f.south west);
                \node (C) at ($(f.south east)!0.5!(f.south west)+(0,-0.3)$) {(e)};
            \end{tikzpicture}
    \caption{Jumping while traversing a horizontal distance.
    Top row:  a sequence of frames from an overhead camera.  Black floor markings indicate distances of one body length. Red lines indicate the landing positions of the robot and the distances traversed between them. The distance indicated in panel (c) is approximately 4 body lengths and the distance indicated in panel (e) is approximately 6 body lengths. 
    Bottom row: corresponding images from the same sequence, taken from a side view panning along with the robot. }
    \label{fig:horjump_fig}
\end{figure*}

\section{Experimental results}\label{sec:experiments}
\subsection{Jumping experiment} \label{sec:vertexperiment}
Following the results of the numerical experiment described in the previous section, the same experiment is conducted on the robot described in Sec. \ref{sec:robot_design}. In this experiment, the same reference velocity is provided to the motor as given by Eq. \ref{eq:vert_jump_dpsi}.  The result of this experiment is shown in Fig. \ref{fig:vert_jump_exp}.  Video of the experiment was recorded at 240 frames/second (fps) from a side view with a stationary camera.  Fig. \ref{fig:vert_jump_traj_pic} shows four frames from this recording overlaid with one another to represent the path followed by the robot during the jump.  The video was post-processed using the software, Tracker (https://physlets.org/tracker/) to obtain the trajectory followed by a point located on the center of the front hoop.  This trajectory is given by the curve plotted over the image and is shown over time in Figs. \ref{fig:vert_jump_xtraj}, \ref{fig:vert_jump_ytraj}.  The four points on the trajectory that are overlaid in Fig. \ref{fig:vert_jump_traj_pic} are indicated the plots in Fig. \ref{fig:vert_jump_xtraj}, \ref{fig:vert_jump_ytraj} by the blue circles. 
It can be seen from this trajectory that the motion of the robot agrees with the numerical simulation of the previous section.  The robot completes a small hop followed by a small jump and then a large jump. The simulation of the previous section predicts the small jump height well, both showing this height to be approximately 0.7-0.8 body lengths.  In the experiment, following this first jump, the robot rebounds into a second jump, achieving a height of approximately 2.4 body lengths, as shown in Fig. \ref{fig:vert_jump_traj_pic}. Fig. \ref{fig:vert_dpsi} shows the relative angular velocity of the pendulum to the hoop during the experiment as measured from the onboard encoder along with the corresponding reference signal in Eq. \ref{eq:vert_jump_dpsi}. Fig. \ref{fig:vert_dphi} shows the wheel angular velocity as measured from the onboard IMU.

\subsection{Horizontal jumping}
In an additional experiment, we apply a different reference angular velocity in an attempt to clear a larger horizontal distance during the jump. In particular, we attempt to have the robot reach a larger horizontal velocity before initiating the jump. This is done by driving the pendulum so that the angular velocity of the wheel increases more rapidly, accelerating from rest to an angular velocity of approximately 23 rad/s over 3.75s. At this point, the jump is initiated by applying an angular velocity reference of 150 rad/s in the same direction as the rotation of the pendulum for jumping.  This reference signal is summarized as follows. 
\begin{equation}\label{eq:hor_jump_dpsi}
\dot{\psi}_\text{ref}(t) =
\begin{cases}
-6.22t & t < 3.75\\
-150 & 3.75 < t < 4.75
\end{cases}
\end{equation}
The direction of rotation for initiating the jump for this test was chosen to be the same direction as the rolling direction because it was seen in previous experiments (including the experiment of Fig. \ref{fig:vert_jump_exp}) that by initiating the jump by swinging the pendulum counter to the direction needed for rolling, while leading to a larger vertical height of the jump, can have a braking effect on the rolling motion, leading to a reduced horizontal velocity of the center of mass at the loss of contact, and therefore a reduced horizontal distance traversed while in flight. 

\begin{figure}
    \centering
        \begin{subfigure}[b]{0.49\linewidth}
        \includegraphics[width = \linewidth]{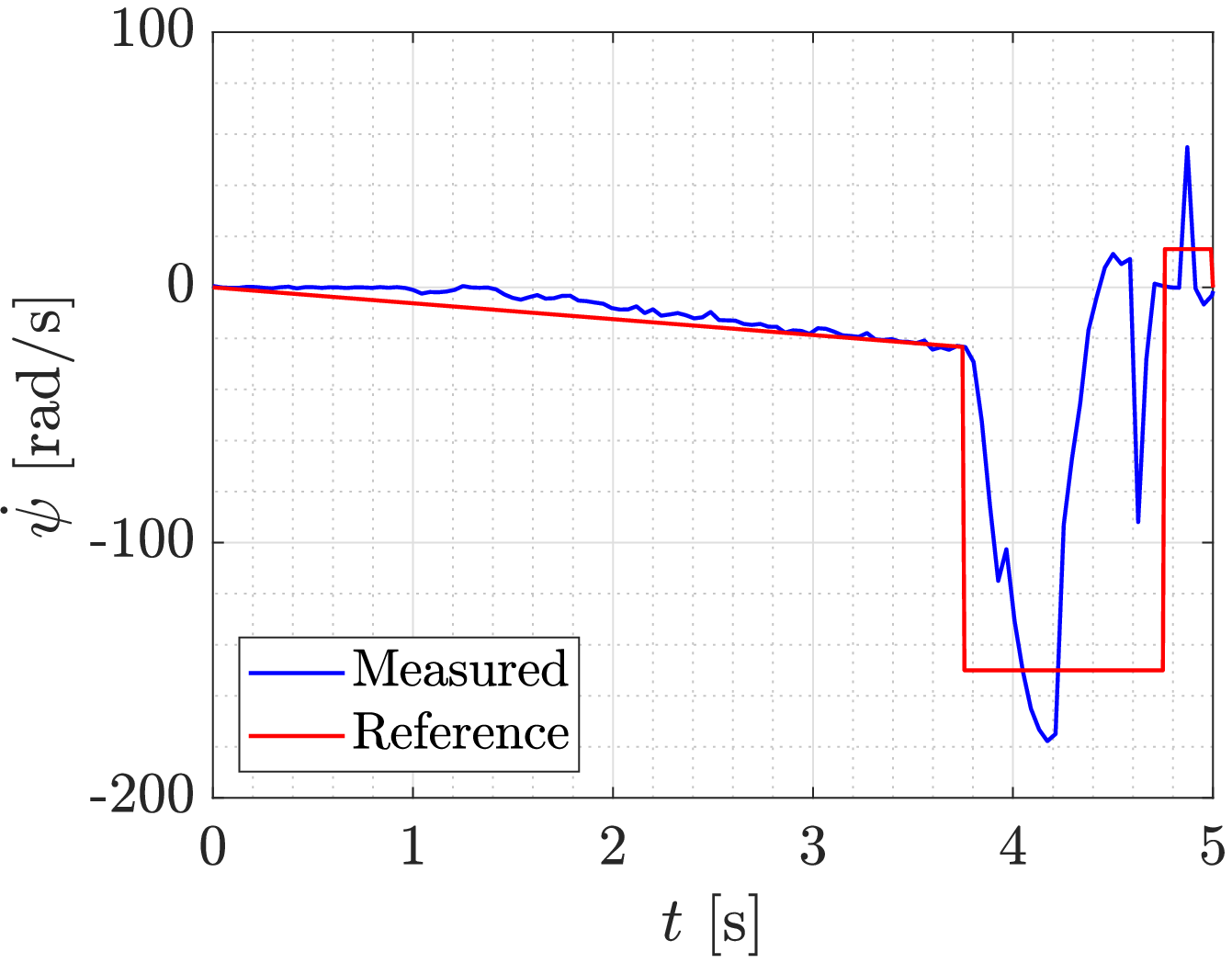}
        \caption{}
        \label{fig:hor_dpsi}
    \end{subfigure}
    \begin{subfigure}[b]{0.49\linewidth}
        \includegraphics[width = \linewidth]{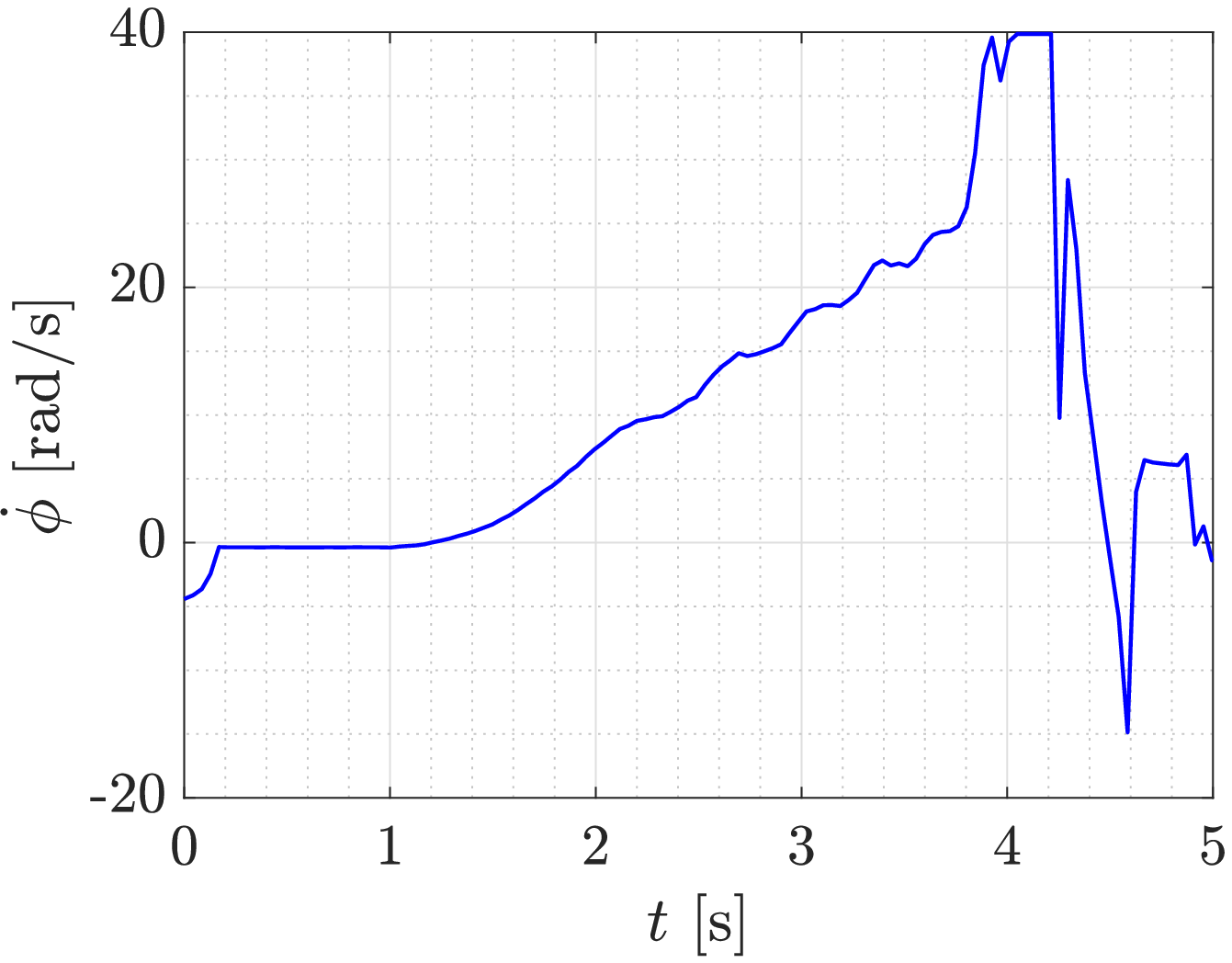}
        \caption{}
        \label{fig:hor_dphi}
    \end{subfigure}
    \caption{Measured data from horizontal jumping experiment shown in Fig. \ref{fig:horjump_fig}. (a) Relative angular velocity between pendulum and wheel. (b) Angular velocity of the wheel.}\vspace{-1ex}
    \label{fig:hor_data}
\end{figure}

Results from this test are shown in Fig. \ref{fig:horjump_fig}. The trial was recorded at 240 fps from a side view taken from a panning camera and additionally from a stationary camera mounted overhead, recording at 120 fps.  Five still images from each of these views are aligned and shown in Fig. \ref{fig:horjump_fig}.  The top row of Fig. \ref{fig:horjump_fig} shows views from the overhead camera, where the black markings on the ground are evenly spaced ticks with spacings of one diameter separating each.  

During this experiment, once the jump is initiated, the robot completes one small hop, followed by two jumps in which a significant horizontal distance is cleared.  The red marking in the top panel of Fig. \ref{fig:horjump_fig}a indicates the landing position after the first small hop.  Fig. \ref{fig:horjump_fig}b shows the robot in mid-flight after launching from the position shown in Fig. \ref{fig:horjump_fig}a.  Fig. \ref{fig:horjump_fig}c shows the landing position of the robot following the first significant jump. The top panel of Fig. \ref{fig:horjump_fig}c shows a red marking indicating the landing position along with a line indicating the distance traversed during this first jump, which spans approximately 4 body lengths. 
Fig. \ref{fig:horjump_fig}d shows the robot in mid-flight after launching from the position shown in Fig. \ref{fig:horjump_fig}c .  Finally, Fig. \ref{fig:horjump_fig}e shows the landing position of the robot during the second jump, with the top panel indicating the distance traversed during this jump, a horizontal span of over 6 body lengths. Fig. \ref{fig:hor_data} shows the measured data collected in this experiment from the onboard encoder and IMU, respectively, along with the reference signal for the relative angular velocity in Eq. \ref{eq:hor_jump_dpsi}.

\section{Conclusion}
We have presented a novel design for a pendulum-driven wheel robot that can both roll and jump. In experiments, the robot was capable of jumping up to 2.4 body lengths vertically and, in separate tests, was able to traverse horizontal lengths of over 6 body lengths in flight. Ongoing and future work on this robot include the development and implementation of trajectory planning and model-based control strategies, consideration of the full dynamics including 3D rotations and collision effects, designs and control strategies to allow for stabilization following a jump. 

\addtolength{\textheight}{-12cm}   

\bibliographystyle{ieeetr}
\bibliography{jumping}

\begin{thebibliography}{10}

\bibitem{msu_jumper_2013}
J.~Zhao, J.~Xu, B.~Gao, F.~Cintron, M.~Mutka, and L.~Xiao, ``Msu jumper: A
  single-motor-actuated miniature steerable jumping robot,'' {\em IEEE
  Transactions on Robotics}, vol.~29, no.~3, pp.~602--614, 2013.

\bibitem{penn_jerboa_2015}
A.~De and D.~Koditschek, ``Parallel composition of templates for tail-energized
  planar hopping,'' in {\em IEEE International Conference on Robotics and
  Automation}, pp.~4562--4569, 2015.

\bibitem{grillo_2007}
U.~Scarfogliero, C.~Stefanini, and P.~Dario, ``Design and development of the
  long-jumping “grillo” minirobot,,'' in {\em IEEE International Conference
  on Robotics and Automation}, pp.~467--472, 2007.

\bibitem{salto_2016}
D.~W. Haldane, M.~M. Plecnik, J.~K. Yim, and R.~S. Fearing, ``Robotic vertical
  jumping agility via series-elastic power modulation,'' {\em Science
  Robotics}, vol.~1, no.~1, 2008.

\bibitem{littlewood}
J.~Littlewood, {\em A Mathematicians Miscellany}.
\newblock London: Methuen \& Co. Ltd, 1953.

\bibitem{tokieda_hopping_1997}
T.~F. Tokieda, ``The {Hopping} {Hoop},'' {\em American Mathematical Monthly},
  vol.~104, pp.~152--154, Feb. 1997.

\bibitem{moffatt_prs_2005}
Y.~Shimomura, M.~Branicki, and H.~Moffatt, ``Dynamics of an axisymmetric body
  spinning on a horizontal surface. {II}. {S}elf-induced jumping,'' {\em
  Proceedings of the Royal Society A}, vol.~461, no.~2058, pp.~1753--1774,
  2005.

\bibitem{ivanov_rcd_2008}
A.~Ivanov, ``On detachment conditions in the problem on the motion of a rigid
  body on a rough plane,'' {\em Regular and Chaotic Dynamics}, vol.~13,
  p.~355–368, 2008.

\bibitem{bronars_prs_2019}
A.~Bronars and O.~M. O'Reilly, ``Gliding motions of a rigid body: the curious
  dynamics of littlewood's rolling hoop,'' {\em Proceedings of the Royal
  Society A}, vol.~475, no.~2231, p.~20190440, 2019.

\bibitem{tallapragada_passive_2019}
P.~Tallapragada, J.~Buzhardt, and R.~Seney, ``A passive jumping mechanism,'' in
  {\em Dynamic Systems and Control Conference}, ASME, 2019.

\bibitem{mcgeer1990passive}
T.~McGeer, ``Passive dynamic walking,'' {\em {International Journal of Robotics
  Research}}, vol.~9, no.~2, pp.~62--82, 1990.

\bibitem{garcia1998simplest}
M.~Garcia, A.~Chatterjee, A.~Ruina, and M.~Coleman, ``{The Simplest Walking
  Model: Stability, Complexity, and Scaling},'' {\em Journal of Biomechanical
  Engineering}, vol.~120, pp.~281--288, 04 1998.

\bibitem{collins2005efficient}
S.~Collins, A.~Ruina, R.~Tedrake, and M.~Wisse, ``Efficient bipedal robots
  based on passive-dynamic walkers,'' {\em Science}, vol.~307, no.~5712,
  pp.~1082--1085, 2005.

\bibitem{armour_bb_2007}
R.~Armour, K.~Paskins, A.~Bowyer, J.~Vincent, and W.~Megill, ``Jumping robots:
  a biomimetic solution to locomotion across rough terrain,'' {\em
  Bioinspiration \& {B}iomimetics}, vol.~2, no.~3, p.~S65, 2007.

\bibitem{shinichi_ICRA_2007}
Y.~Matsuyama and S.~Hirai, ``Analysis of circular robot jumping by body
  deformation,'' in {\em International Conference on Robotics and Automation
  (ICRA)}, IEEE, 2007.

\bibitem{misu_iros_2018}
K.~Misu, A.~Yoshii, and H.~Mochiyama, ``A {Compact} {Wheeled} {Robot} that
  {Can} {Jump} while {Rolling},'' in {\em {International} {Conference} on
  {Intelligent} {Robots} and {Systems} ({IROS})}, IEEE, Oct. 2018.

\bibitem{hayashi_high-performance_2001}
R.~Hayashi and S.~Tsujio, ``High-performance jumping movements by pendulum-type
  jumping machines,'' in {\em {International} {Conference} on {Intelligent}
  {Robots} and {Systems}. ({IROS})}, IEEE, 2001.

\bibitem{hockman2017design}
B.~J. Hockman, A.~Frick, R.~G. Reid, I.~A. Nesnas, and M.~Pavone, ``Design,
  control, and experimentation of internally-actuated rovers for the
  exploration of low-gravity planetary bodies,'' {\em Journal of Field
  Robotics}, vol.~34, no.~1, pp.~5--24, 2017.

\bibitem{ho2017mascot}
T.-M. Ho, V.~Baturkin, C.~Grimm, J.~T. Grundmann, C.~Hobbie, E.~Ksenik,
  C.~Lange, K.~Sasaki, M.~Schlotterer, M.~Talapina, {\em et~al.},
  ``Mascot—the mobile asteroid surface scout onboard the hayabusa2 mission,''
  {\em Space Science Reviews}, vol.~208, pp.~339--374, 2017.

\bibitem{geist_wheelbot_2022}
A.~R. Geist, J.~Fiene, N.~Tashiro, Z.~Jia, and S.~Trimpe, ``The {Wheelbot}: {A}
  {Jumping} {Reaction} {Wheel} {Unicycle},'' {\em IEEE Robotics and Automation
  Letters}, vol.~7, pp.~9683--9690, Oct. 2022.

\bibitem{romanishin20153d}
J.~W. Romanishin, K.~Gilpin, S.~Claici, and D.~Rus, ``3d m-blocks:
  Self-reconfiguring robots capable of locomotion via pivoting in three
  dimensions,'' in {\em International Conference on Robotics and Automation
  (ICRA)}, IEEE, 2015.

\bibitem{muehlebach2016cubli}
M.~Muehlebach and R.~D’Andrea, ``Nonlinear analysis and control of a
  reaction-wheel-based 3-d inverted pendulum,'' {\em IEEE Transactions on
  Control Systems Technology}, vol.~25, no.~1, pp.~235--246, 2016.

\bibitem{kayacan2012sphere}
E.~Kayacan, Z.~Y. Bayraktaroglu, and W.~Saeys, ``Modeling and control of a
  spherical rolling robot: a decoupled dynamics approach,'' {\em Robotica},
  vol.~30, no.~4, pp.~671--680, 2012.

\bibitem{howell2000torsobiped}
G.~W. Howell, {\em Analysis and control of superarticulated biped robots}.
\newblock PhD thesis, Boston University, 2000.

\bibitem{sanchez_differential_2020}
S.~Sanchez and P.~A. Bhounsule, ``A differential drive rimless wheel that can
  move straight and turn,'' in {\em {International} {Conference} on {Advanced}
  {Intelligent} {Mechatronics} ({AIM})}, IEEE, 2020.

\bibitem{taylor_dynamics_2010}
A.~Taylor and M.~Fehrs, ``The dynamics of an eccentrically loaded hoop,'' {\em
  American Journal of Physics}, vol.~78, pp.~496--498, May 2010.

\end{thebibliography}
\end{document}